# Agentic AI Ecosystems in Higher Education: A Perspective on AI Agents to Emerging Inclusive, Agentic Multi-Agent AI Framework for Learning, Teaching and Institutional Intelligence

**Vidya K Sudarshan[1,2*], Anushka Sisodia[1], Reshma A Ramachandra[2], Batra Sia[1], Josephine Chong Leng Leng[1]**

**[1]College of Computing and Data Science, Nanyang Technological University, NTU, Singapore**

**[2]DeepMed Ptd Ltd, India**

***Corresponding Author Email:** vidya.sudarshan@ntu.edu.sg

## Abstract

Integration of artificial intelligent (AI) agents in higher education is transforming teaching, learning and administrative processes. Although existing AI agents effectively support individual tasks, their implementation remains fragmented and inefficient for handling the complexity of educational institutions. This highlights a significant research gap: the lack of integrated eco-system-level agentic multi-agent AI platform capable of coordinated planning, reasoning, and adaptive decision-making across multiple educational functions. This paper presents a forward-looking perspective on agentic multi-agent AI platform in higher education, consisting interconnected autonomous, goal driven agents that support learning, teaching, and institutional operations. It addresses timely and critical questions: Can agentic AI represent the next generation of intelligent systems in tertiary education? Can they collectively support seamless coordinated operations across teaching, learning and administrative support? To what extent can such systems foster inclusive and equitable learning for diverse learners with special educational needs? To ground this perspective, a thematic analysis of existing literature identifies four dominant themes: task-specific fragmented AI tools, the transition from single-agent to multi-agent systems, limited cross-functional integration, and insufficient focus on inclusivity and accessibility. Findings reveal a clear gap between current AI implementations and the needs of holistic, learner-centered educational ecosystem. The paper synthesizes challenges and outlines future research directions for scalable human-aligned, and inclusive agentic AI platform. The significant contribution is the incorporation of inclusive learning perspectives, highlighting how coordinated agentic multi-agent platform can support diverse learners through adaptive, multimodal interventions.

## 1. Introduction

Over the decades, adoption of artificial intelligence (AI) into tertiary education has notably expedited and the progress in generative AI (Gen AI) and large language models (LLM) have led to the rise of autonomous AI agents capable of reasoning, learning, and performing complex tasks (Russel et al., 2020). An AI agent is a system or a software algorithm capable of acting autonomously to understand user request, plan and execute the tasks. The agent utilizes LLM as their core reasoning engine which helps to interface with its environment, other models and other aspects of a system or network as needed to achieve specific goals (Xi et al., 2023). These agents can range from simple task-specific applications (e.g., automated grading, intelligent tutoring, and learning analytics) to highly advanced interactive, adaptive and context-aware learning system (Russel et al., 2020; Kasneci et al., 2023; Xi et al., 2023). These significant transition from rule-based expert systems to more complex AI agents have demonstrated considerable benefits in improving learning outcomes, enhancing engagement, and supporting data-driven decision-making (Holmes et al., 2019; Zawacki-Richter et al., 2019; Kostopoulos, G et al., 2025). Thus, AI agents have evolved from single-purpose tools to more generalizable, multi-agent frameworks capable of collaborative decision-making and adaptive interaction (Wooldridge 2009). However, their deployment in higher education still remails largely fragmented, focusing individual stakeholder and addressing isolated functions, rather than running as integrated platform (ecosystem). This isolated, non-integrated working of multi-agents limits the effectiveness to the complex and interdependent and / or interconnected tasks requirements of modern educational systems such as teaching, learning, and other administrative processes. This uncovers a significant gap like lack of coordinated, interconnected multi-agents' platform in the current educational environments capable of executing multiple functions by integrating requests from multiple stakeholders including students with diverse educational needs. Based on this research gap, this paper argues that the future of AI in tertiary education is pointing toward a new paradigm, agentic AI, and development of agentic multi-agents' platform or ecosystem where multiple specialized agents collaborate across learning, teaching, and institutional domains to create a cohesive, adaptive, and intelligent environment (Wooldridge, 2009; Russell et al., 2020). In addition, this paper extends beyond efficiency and automation to highlight the perceptive on inclusive and equitable learning, recommending how agentic AI platform can contribute to support diverse learners through coordinated, personalized, and context-aware interventions. This paper presents four main contributions. First, we present conceptualised transition from isolated AI agents to agentic multi-agent platforms in tertiary education, by providing organised perceptive for understanding the evolution of AI-enabled educational systems. Second, we propose single agentic multi-agent ecosystem or platform, a unified multi-stakeholder framework, for serving students, educators and institutional processes. Third, we introduce the perception of inclusive agentic AI platform and its potential in supporting students with SENs through coordinated multi-agent interactions. Finally, we highlight critical challenges and outline future research directions for building human-aligned, scalable, and inclusive agentic AI platform in tertiary education.

## 2. From AI Agents to Agentic Ecosystems

The introduction of AI agents has marked a noteworthy move toward autonomous systems capable of responding to queries, recommending resources or supporting basic instructional interactions with autonomy and interactivity, enabling more responsive and personalized experiences (Wooldridge 2009; Luckin et al., 2016; Luckin 2017; Russel et al., 2020; Chen et al., 2020; Xi et al., 2023; Wang S et al., 2024). Figure 1 shows a block diagram of a typical AI agent.

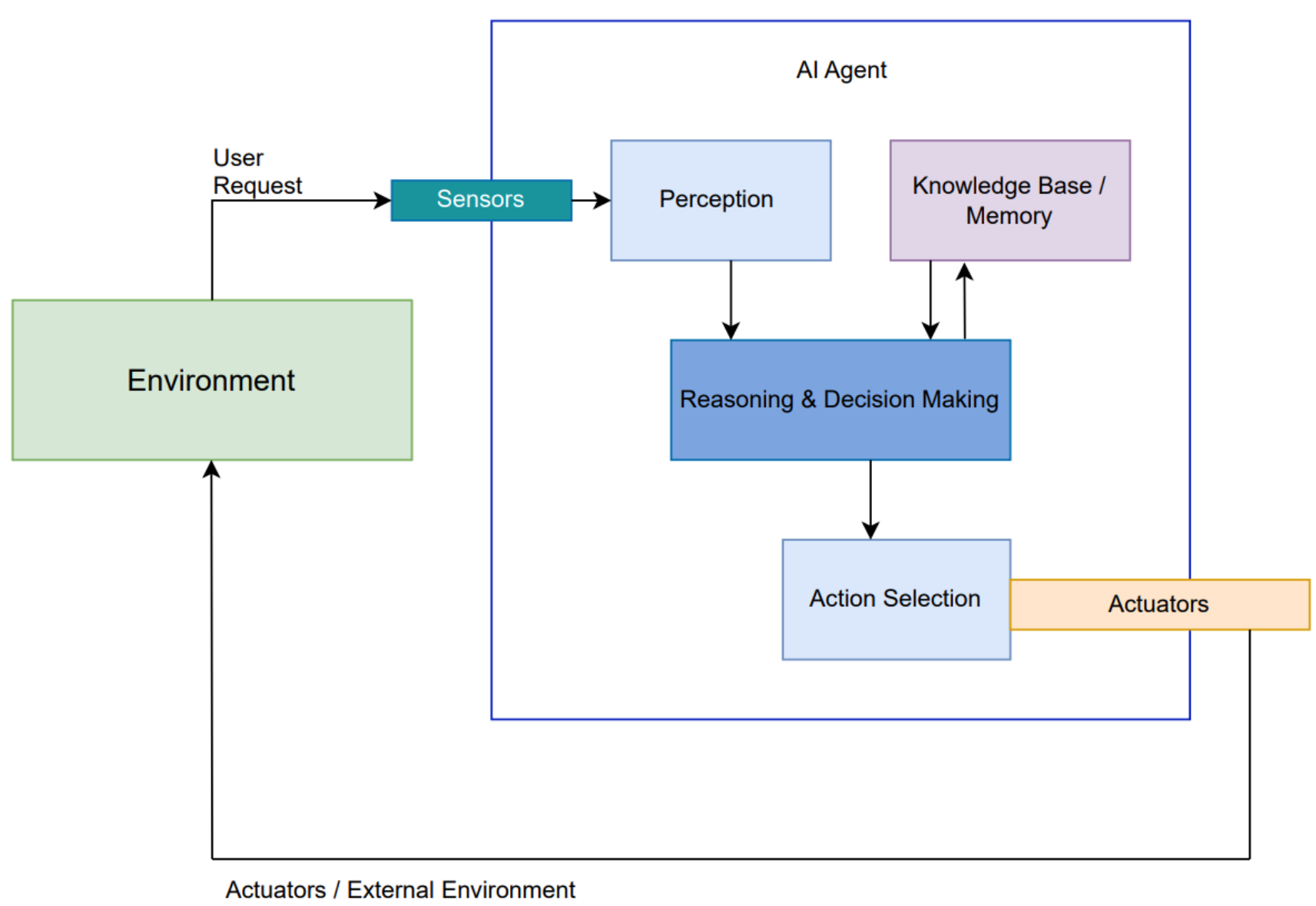

**Figure 1: Block Diagram of a Typical AI agent**

A basic AI agent functions through a perception–decision–action loop incorporating input, internal reasoning mechanism, and output. An AI agent is made up of a few key parts that work together to sense the environment, think about task, and act. First, the perception module collects information—like reading text, hearing speech, or receiving data—and turns it into something the agent can understand. Next, the memory helps the agent keep track of important details. Short-term memory holds information for the current task (like the last message in a conversation), while long-term memory stores useful knowledge the agent can use again later. The reasoning or cognitive module is the part that thinks to interpret the inputs and makes decisions about the best actions to take as the next step. Once the decision is made, the action module carries it out—such as sending a message, generating content, calling a tool, or interacting with external systems. Finally, a simple feedback loop helps the agent learn by checking whether its action worked and adjusting future behaviour. Together, these components allow an AI agent to sense, think, act, and improve over time.

### 2.1 Evolution of AI in Education

The evolution of AI in education spans over six decades (Figure 2), beginning in the 1960s–1970s, a period that witnessed the emergence of foundational computer-assisted instruction systems.

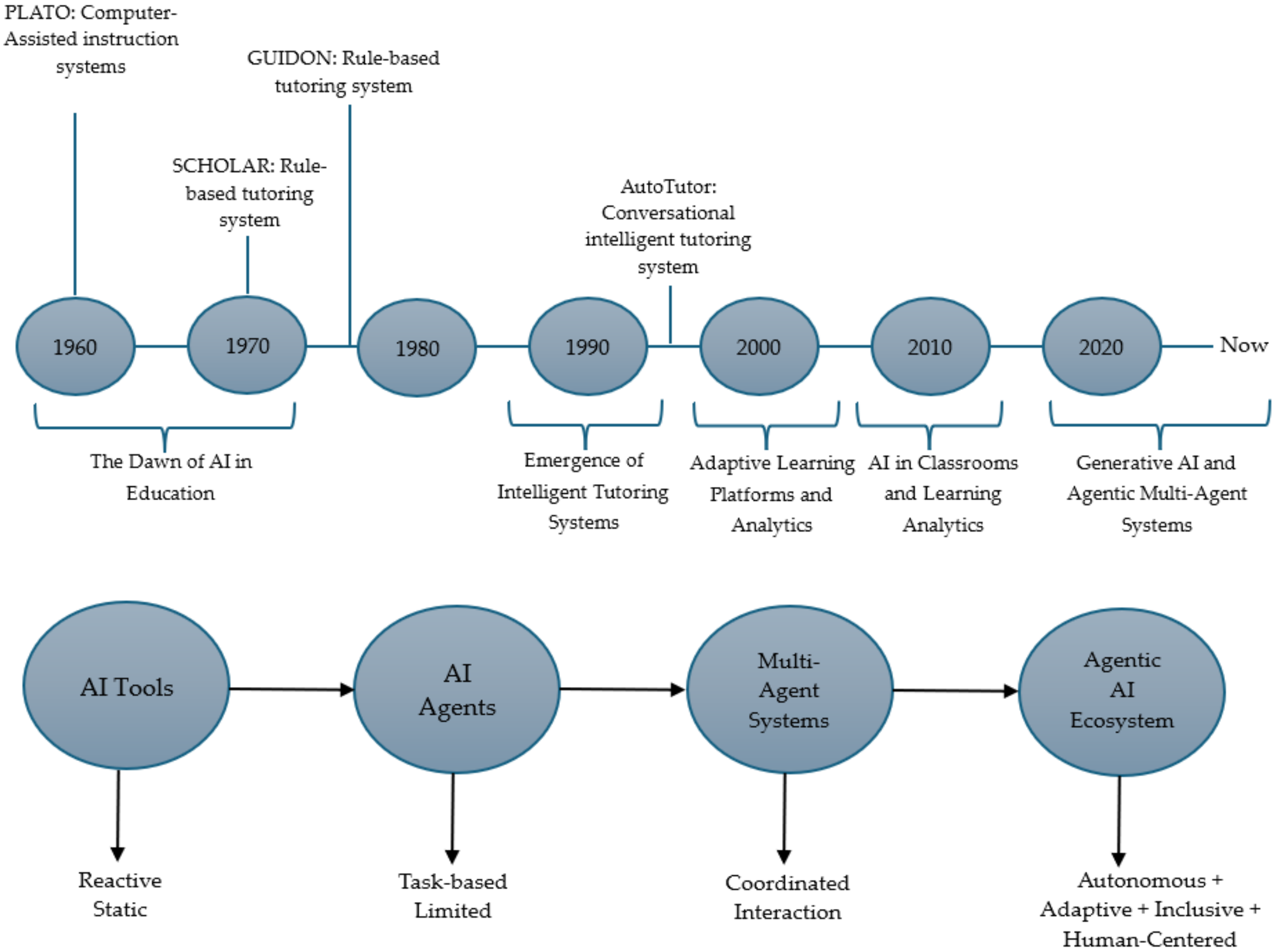

**Figure 2. Integrated view of the evolution of AI in higher education, combining historical development with conceptual progression. The figure illustrates the transition from early computer-assisted instruction systems to intelligent tutoring systems, followed by adaptive learning platforms and generative AI, culminating in agentic AI ecosystems. This progression**

**reflects a shift from reactive, task-specific tools toward autonomous, adaptive, and inclusive multi-agent systems capable of coordinated, ecosystem-level intelligence.**

In the 2020s–Present, generative AI has catalysed a new wave of intelligent agents. This historical perspective provides critical context for understanding the rapid advancements in educational AI—from early rule-based tutoring systems to the rise of increasingly complex, autonomous, and interconnected agentic multi-agent systems that define the current frontier in personalized and intelligent learning environments (Zawacki-Richter et al., 2019; Kostopoulos, G et al., 2025). This trajectory illustrates a clear shift from isolated tools toward increasingly personalized, intelligent, integrated systems capable of supporting dynamic and multifaceted educational processes (Holmes et al., 2019; Zawacki-Richter et al., 2019; Istrate O 2024; Ravaglia 2025). Initial deployments of AI agents in education reported as designed as single-purpose tools, performing narrow tasks such as grading, scheduling, or recommendation. For students, AI agents now offer just-in-time feedback, emotionally responsive tutoring, and adaptive content tailored to individual learning needs. Educators benefit from AI tools that generate lesson plans, provide formative assessment insights, automate grading, and track classroom engagement. Administrators, in turn, leverage intelligent systems for managing institutional workflows, monitoring academic performance at scale, and supporting data-driven decision-making. As AI technologies have matured, the role of agents has expanded significantly beyond one-on-one tutoring to encompass role-specific functions marking a transformative shift in the way AI can be meaningfully integrated into tertiary education (Luckin et al., 2016; Chen et al., 2020; Wang S et al., 2024). These systems typically operate in isolation and offer limited adaptability to changing learner contexts. While effective in specific domains, their lack of integration restricts their ability to address the broader complexity of educational environments (Luckin et al., 2016).

Thus arrived the stage of multi-agent systems (MAS) where multiple agents interact, coordinate, and collaborate to achieve shared goals (Zawacki-Richter et al., 2019; Kostopoulos, G et al., 2025). In educational settings, this may involve integration and coordination between tutoring agents, assessment systems, and learning analytics platforms working in tandem. Such systems support distributed intelligence and allow for more sophisticated decision-making processes. As systems matured and institutions started adopting more digital infrastructure, these AI agents began to interoperate forming multi-AI agent configurations or environments, where several agents (e.g., enrolment chatbots, scheduling bots, and analytics dashboards) worked simultaneously but often in parallel, each handling different aspects of tasks in educational institute (Uchoa AP et al., 2025). However, despite these advancements, many multi-agent implementations remain loosely coupled, lacking deep integration and cohesive coordination across institutional functions (Jennings, 1998; Wooldridge, 2009).

**2.2 Agentic AI and Ecosystem-Level Intelligence**

The integration of AI in education has progressed through several distinct stages—beginning with single AI agent, advancing to multiple coordinated agents, and now evolving toward agentic AI platforms, characterized by autonomy, reasoning, goal-directed behaviour, and context-awareness. This marks a significant shift where AI agents not only respond to queries but also proactively analyse data, generate insights, and make context-driven decisions across the university (UUK et al., 2023). Building on this when extended to interconnected agentic multi-agent configurations, leads to the concept of agentic AI ecosystems incorporated with numerous autonomous specialized agents, that can communicate, coordinate tasks, and optimise workflows across different domains and functions. These ecosystems enable coordinated, system-level intelligence, where interactions between agents lead to emergent capabilities that surpass those of individual components.

Unlike traditional agents, agentic systems are not limited to predefined tasks but can continuously plan, learn, and adjust their behavior based on evolving contexts (Wooldridge, 2009; Russell et al., 2020; Derouiche et al., 2025; Nisa et al., 2025). Agentic ecosystems are defined by several key characteristics such as, autonomy (agents operate independently while aligning with system-level goals), coordination (agents communicate and collaborate effectively), adaptability (systems respond dynamically to changing contexts), and context-awareness (decisions are informed by real-time data and user needs) (Russell et al., 2020). In higher education, such ecosystems facilitate continuous and personalised learning support, enable real-time feedback and instructional adaptation, and enhance data-driven institutional decision-making (Zhao A et al., 2024a; Zhao P et al., 2024b). By moving beyond isolated applications toward coordinated intelligence, agentic ecosystems provide a scalable and flexible framework for addressing the complexity of modern educational environments. Figure 3 shows the block diagram of a typical agentic AI platform.

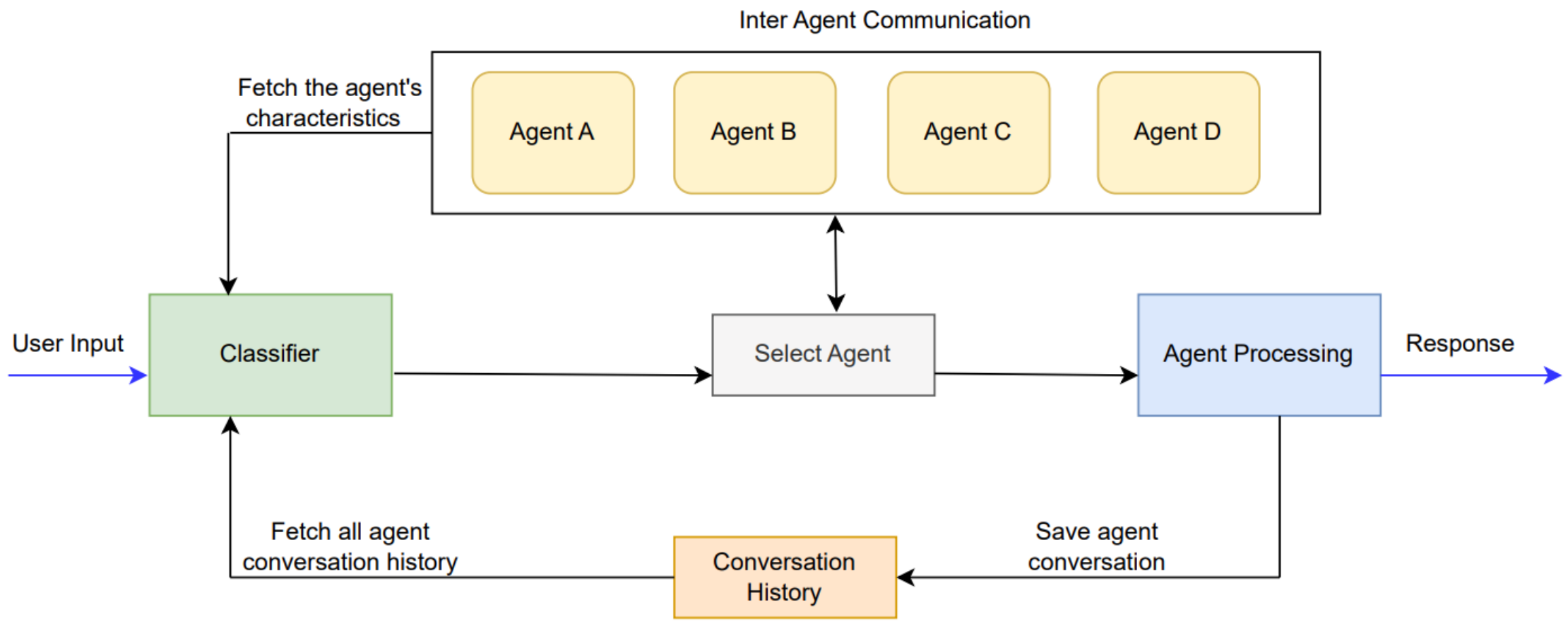

**Figure 3: Block Diagram of a Typical Agentic Multi-Agent AI Platform**

## 3. A Unified Multi-Stakeholder Agentic Framework

A central contribution of our paper is the proposal of a unified multi-stakeholder agentic framework that can serve students, educators, and institutional systems within a single, coordinated AI ecosystem (Castañeda et al., 2018). Unlike traditional approaches that treat each stakeholder in isolation, this framework conceptualizes higher education as a socio-technical system, where multiple agents interact across domains to support learning, teaching, and decision-making processes holistically (Holmes et al., 2019; Luckin et al., 2016). The framework emphasizes not only functional specialization but also inter-agent coordination, enabling continuous feedback loops and adaptive responses across different layers of the educational environment. Figure 4 presents the conceptual structure of the agentic AI ecosystem, highlighting the interactions between key stakeholders and agent types. The architectural realization of this framework is shown in Figure 5, which outlines the layered structure of user interfaces, agent coordination, and underlying data infrastructure.

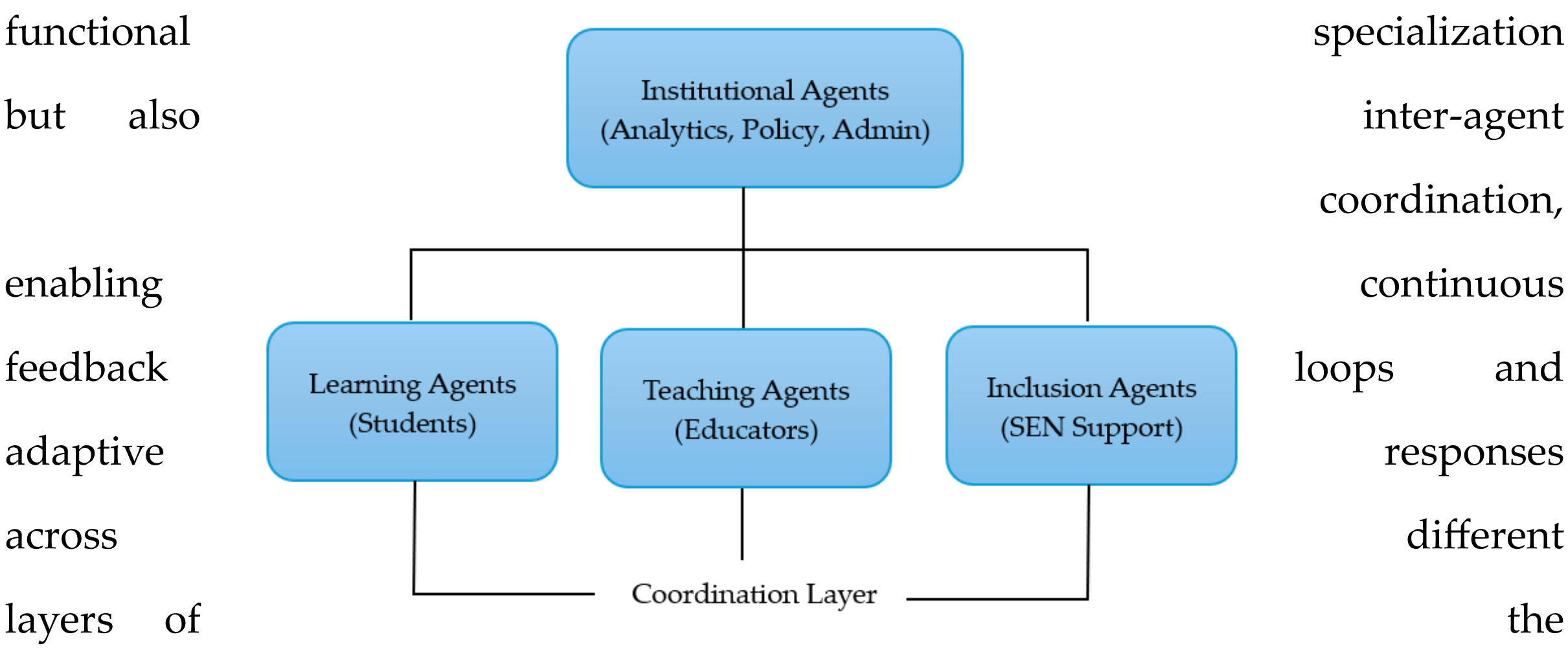

**Figure 4: A unified conceptual multi-stakeholder agentic AI framework illustrating coordinated interactions among learning, teaching, institutional, and inclusion-focused agents within higher education ecosystems.**

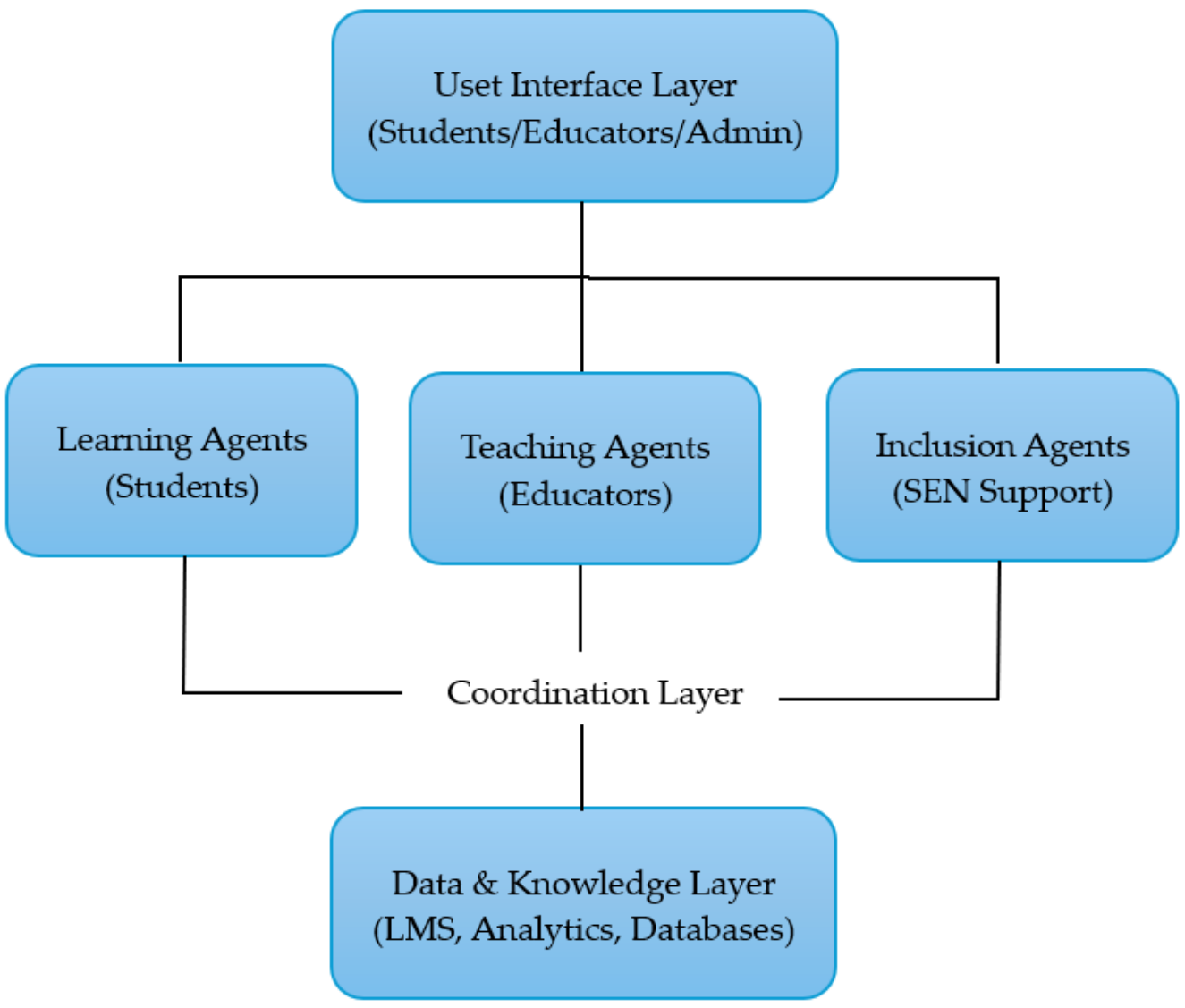

**Figure 5: Layered architecture of an agentic AI ecosystem in higher education, highlighting interactions between user interfaces, multi-agent coordination layers, and underlying data and knowledge infrastructure.**

As illustrated in Figure 4, the proposed multi-stakeholder agentic framework conceptualizes the interaction among learning, teaching, inclusion, and institutional agents within a coordinated educational ecosystem. The architectural realization of this framework is presented in Figure 5, which depicts a layered multi-agent system enabling seamless interaction between users, agents, and underlying data infrastructure. Together, Figures 4 and 5 highlight how the proposed model integrates user interaction, agent-level intelligence, and data-driven processes to support scalable, adaptive, and context-aware educational environments.

### 3.1 Learning Agents

Learning agents are designed to support students and increasingly enabling personalized learning experiences by dynamically customizing educational content to match each student's knowledge level, learning style, emotional state, and interests. These agents support students with coursework, self-study, and making personalised learning pathways that adapt to each learner's performance and progress by providing real-time academic support, such as answering subject matter questions, generating personalised schedules, sending timely study reminders, and recommending resources enhanced by sentiment analysis and affective adaptation (Baillifard A et al., 2023; Baradari et al., 2025). In addition, agents can also serve as creative companions, helping students brainstorm ideas, organize their thoughts, and develop clear outlines for essays and projects. Such capabilities align with established research on intelligent tutoring systems and adaptive learning environments, which demonstrate improvements in learning outcomes and engagement through personalization (Luckin et al., 2016; Holmes et al., 2019). Figure 6 illustrates the workflow of a typical learning agent for students.

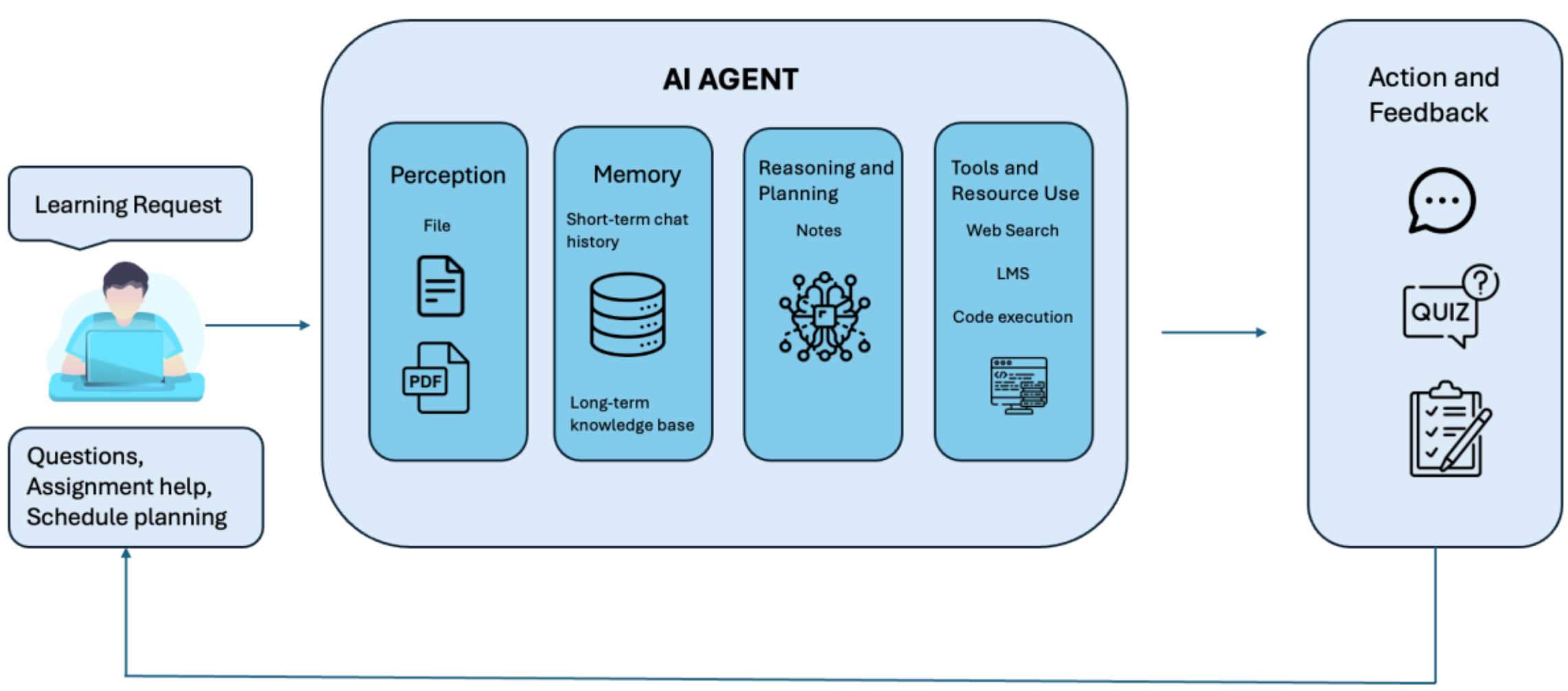

**Figure 6: Illustration of a Typical Learning Agent workflow for Students**

In the context of agentic ecosystems, learning agents extend beyond static personalization by continuously adapting to evolving learner profiles, supporting diverse cognitive needs and promoting self-regulated learning (Cheng et al., 2024; Lisa AD et al., 2024).

### 3.2 Teaching Agents

Figure 7 illustrates the workflow of a typical teaching agent for educators.

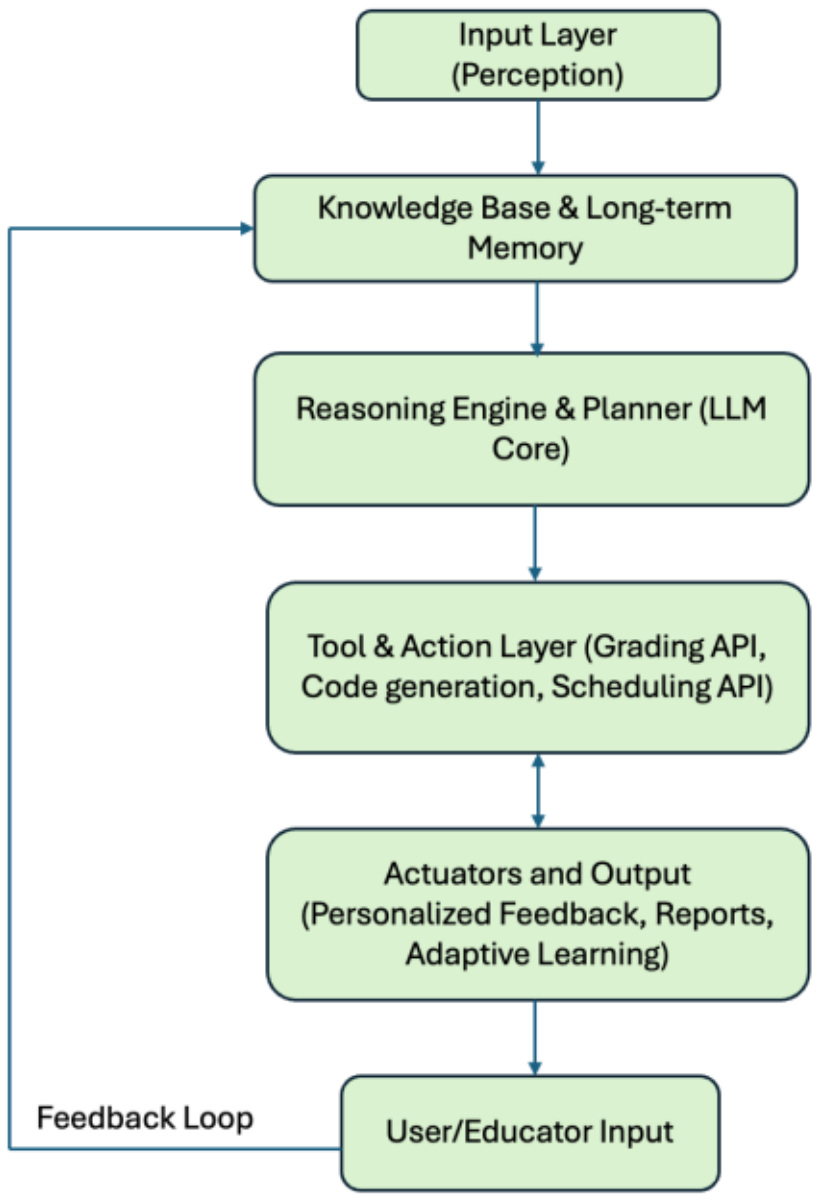

**Figure 7: Illustration of a Typical Teaching Agents for Educators**

Teaching agents function primarily as teaching aides for educators by extending instructional design and pedagogical practices. They support educators to enhance teaching efficiency and personalize student engagement by answering student queries, recommending learning resources, preparing instructional materials, content creation generating lesson plans, quiz questions, video transcripts, automating formative and summative feedback, assisting in assessment processes, providing actionable insights into student performance and automating routine instructional tasks (McKay et al., 2013; Luckin et al., 2022; Sajja R et al., 2024). These capabilities reduce administrative burden while enabling educators to focus more on higher-order teaching activities such as mentoring, critical discussion, and curriculum design (Piech et al., 2015; Naharuddin et al., 2025). Latest developments in generative AI further augment role of teaching agents by enabling context-aware content generation and interactive instructional support (Kasneci et al., 2023; Debora WW et al., 2023; Perkins et al., 2024). Importantly, these agents function as assistive systems rather than replacements, reinforcing the role of educators within a human-centered learning ecosystem.

### 3.3 Institutional Agents

Institutional agents are increasingly used in streamlining institutional processes and enhancing strategic decision-making. Figure 8 illustrates the workflow of a typical institutional agents for administrative tasks.

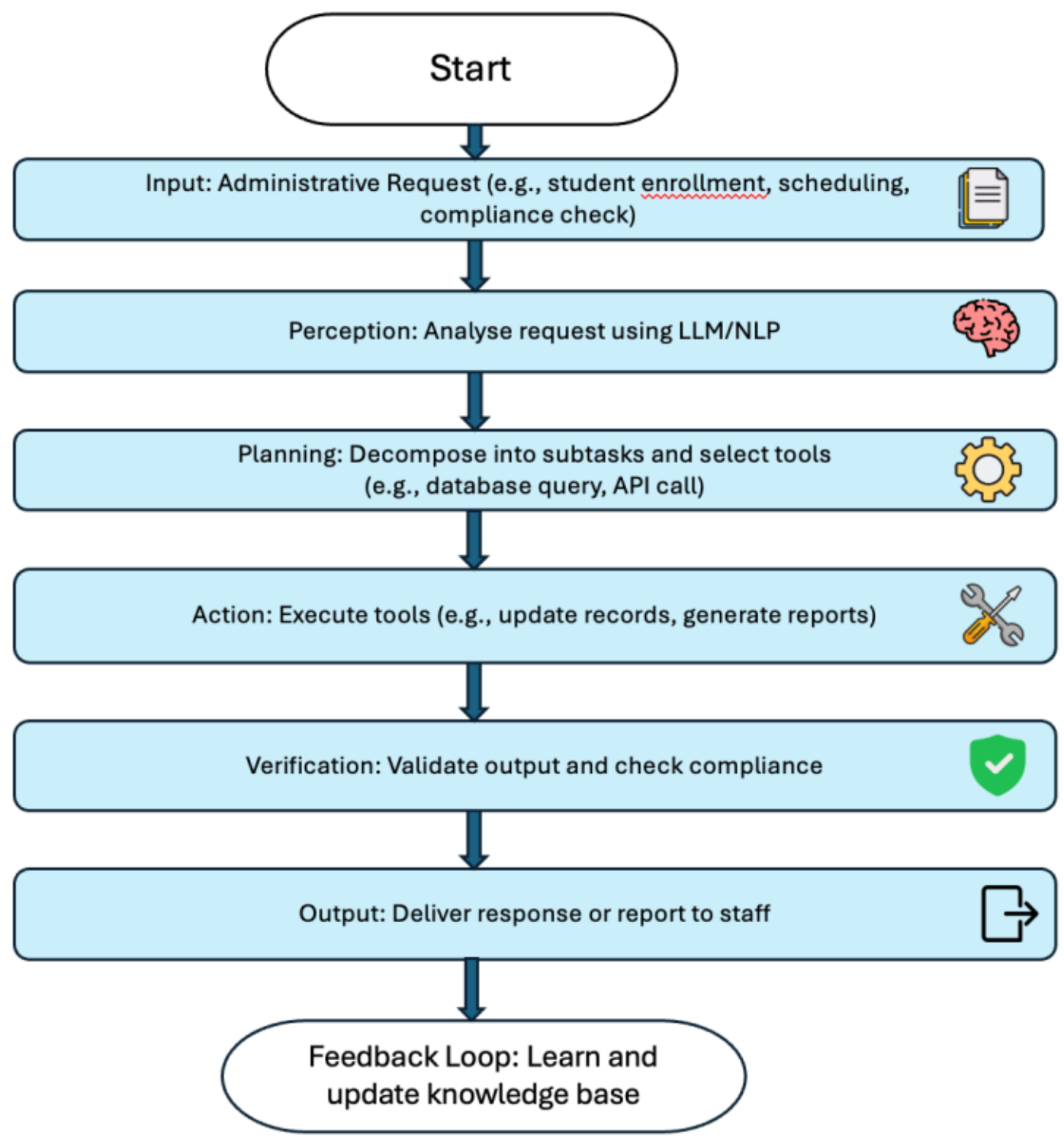

**Figure 8: Illustration of a Typical Institutional Agent for Administrative Tasks**

They operate at the administrative and strategic level, supporting resource optimization, student retention, academic planning, and policy decision-making. Predictive analytics agents analyse student performance data to identify at-risk learners, forecast graduation rates, and assess engagement levels (guide targeted interventions) (Arnold et al., 2012). Enrolment and advising agents assist with managing routine student queries related to registration, financial aid, campus services, course registration, curriculum advising, and student policy navigation. Administrative agents support help to automate onboarding, provide real-time assistance for administrative queries, and enhance student communication throughout their academic journey. Resource scheduling agents optimize the allocation of physical and human resources by intelligently managing classroom bookings, faculty timetables, and exam schedules. These systems contribute to the efficient operation of university infrastructure and reduce administrative friction (Skedda 2024; Robinson 2022). These agents by using large-scale data analytics, provide that inform institutional strategies and improve operational efficiency. Such systems build upon prior work in learning analytics and educational data mining, which highlight the potential of data-driven approaches to enhance institutional decision-making and student success (Zawacki-Richter et al., 2019). This data-driven approach enables proactive intervention and institutional planning (Arnold et al., 2012). These agents are scalable and reduce the burden on human advisors by handling high-volume FAQs and routine queries efficiently (Page et al., 2016; Goel et al., 2016a; Goel et al., 2016b; Georgia State University 2020; Mainstay 2023; Maiti et al., 2024).

Within an agentic ecosystem, institutional agents are not isolated dashboards but active participants in a coordinated system, interacting with learning and teaching agents to align operational decisions with pedagogical goals.

### 3.4 Cross-Functional Coordination

The defining feature of agentic ecosystems lies in the dynamic interaction and coordination between agents across different domains. Unlike traditional systems that operate in silos, agentic frameworks enable (i) feedback loops between learning and teaching processes, (ii) alignment between pedagogical practices and institutional objectives, (iii) real-time adaptation based on learner behavior and contextual data, and (iv) holistic system optimization through distributed intelligence. This coordination reflects principles from multi-agent systems theory, where collaboration among agents leads to emergent system-level intelligence (Jennings et al., 1998; Wooldridge, 2009). In educational contexts, such interactions enable a shift from fragmented AI applications to, integrated, adaptive ecosystems capable of supporting complex and diverse learning environments. Figure 9 illustrates the evolutionary progression of AI systems in higher education, from isolated single-agent setups to interconnected unified multi-stakeholder agents and fully agentic multi-agent workflows. Figure 10 shows the architecture of a possible agentic multi-agent AI platform in education.

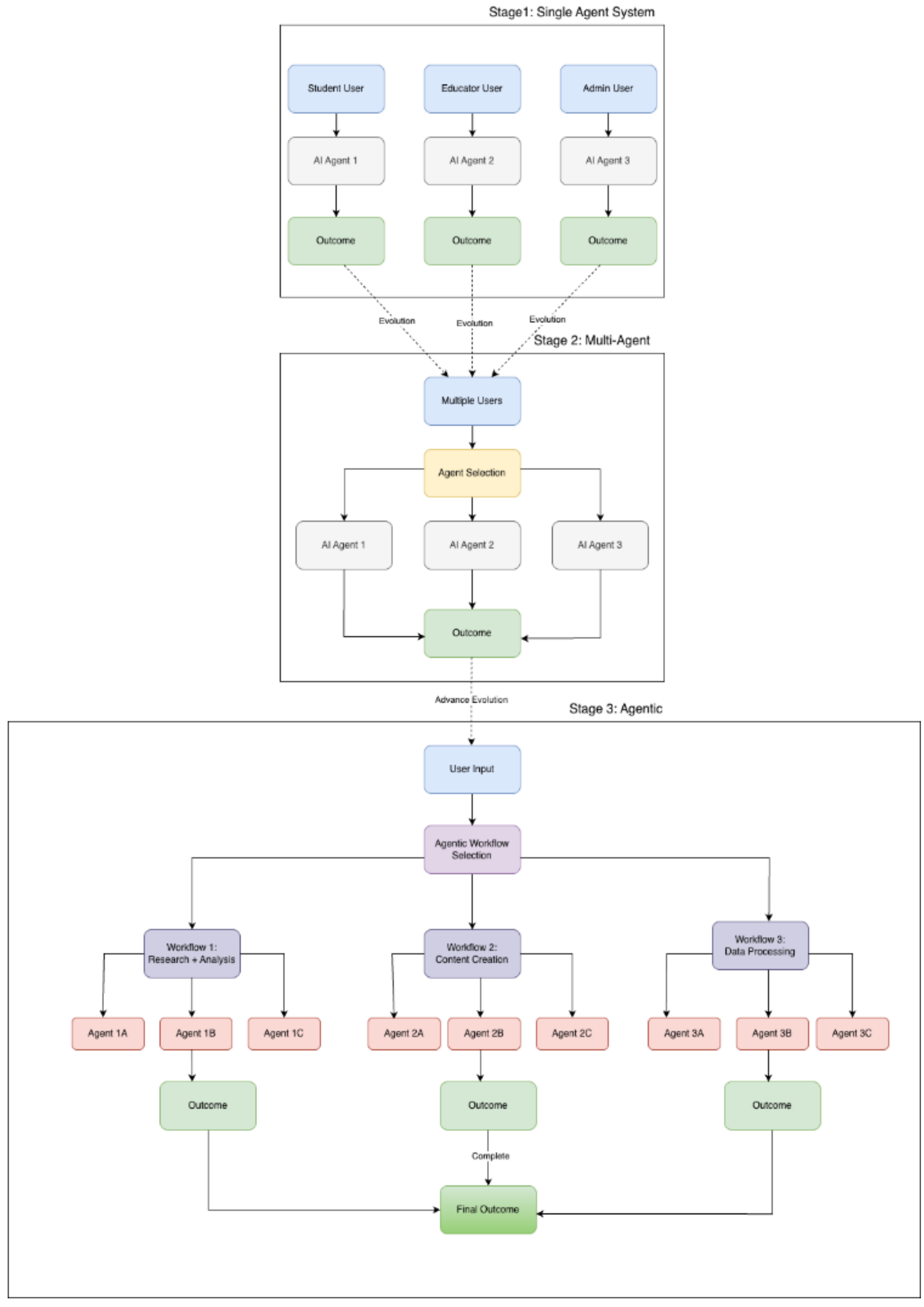

**Figure 9: Evolution from Single-Agent to Unified Multi-Stakeholder Agentic Multi-Agent Systems**

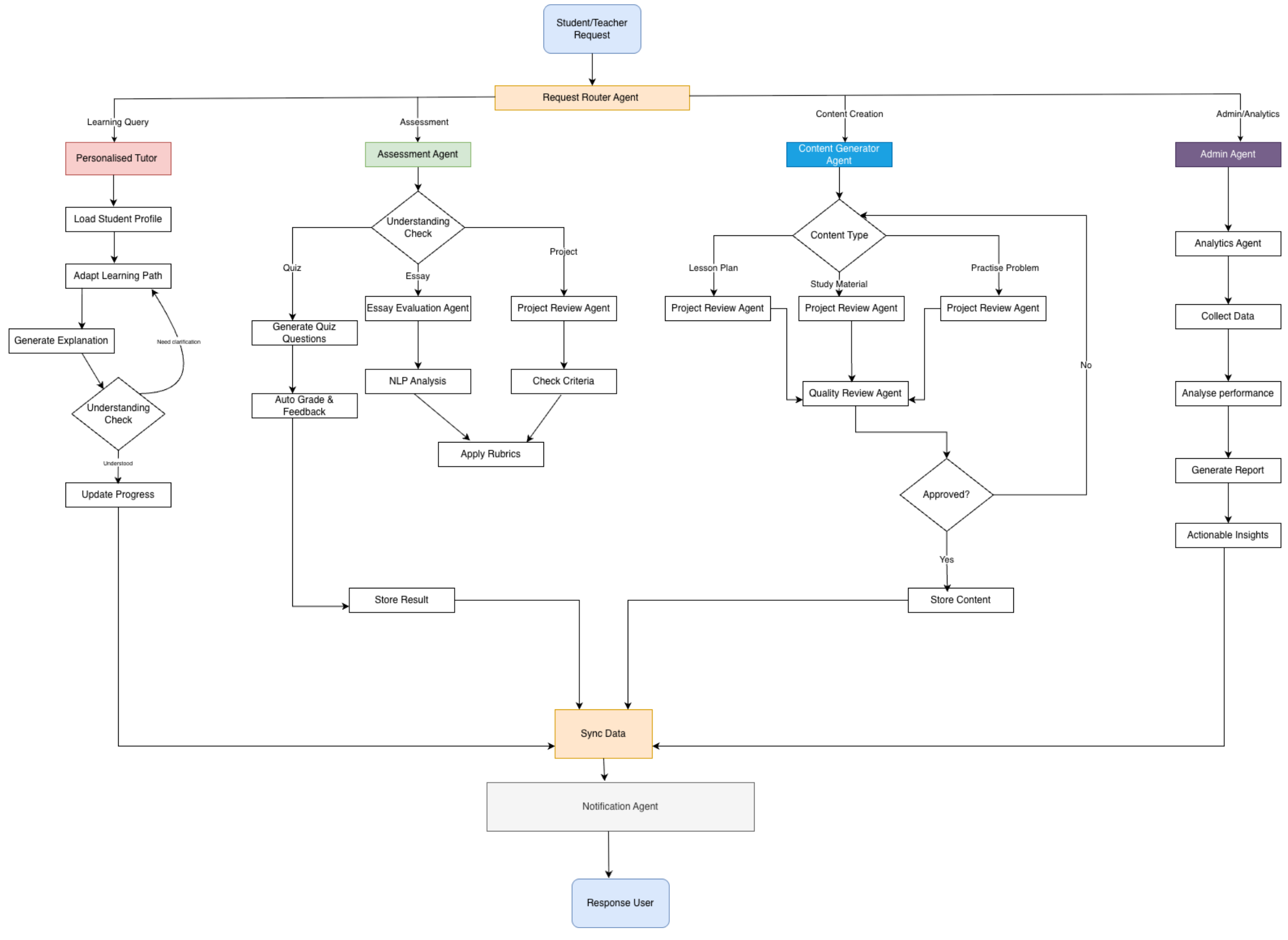

**Figure 10: Architecture of an agentic multi-agent AI platform in education**

## 4. Agentic AI for Inclusive and Equitable Learning

Higher education institutions are facing increasing demands to provide personalized, inclusive, and equitable learning experiences. Students today represent a diverse population with varying cognitive abilities, learning preferences, cultural backgrounds, and accessibility needs. In particular, students with SENs, including those with learning disabilities, neurodiversity, and sensory impairments, require adaptive and supportive learning environments that traditional systems often fail to provide at scale. Thus, one of the most transformative aspects of agentic AI ecosystems is their potential to operationalize inclusive pedagogy by addressing the needs of diverse learners within higher education. Inclusive pedagogy emphasizes the need to design learning environments that accommodate variability in learners' cognitive, sensory, and socio-emotional characteristics, ensuring equitable participation and access for all students (Florian et al., 2011; Luckin et al., 2016). While existing AI-driven educational technologies have demonstrated promise in personalization, their ability to support inclusion at scale remains limited due to fragmentation and lack of coordination (Holmes et al., 2022). Agentic AI ecosystems extend this capability by enabling coordinated, multi-agent support systems that dynamically adapt to individual learner needs. Through continuous interaction, context awareness, and real-time feedback, these systems offer a pathway toward equitable personalization, where inclusivity is embedded within the architecture rather than treated as an add-on feature.

### 4.1 Inclusive Agentic AI Ecosystems

AI has shown significant potential in supporting accessibility through adaptive and assistive technologies including intelligent tutoring systems, multimodal interfaces, and personalized learning environments (Luckin et al., 2016). Building on this foundation, an inclusive agentic AI ecosystem is one which works as a coordinated network of agents that adapts to diverse learner profiles, provides multimodal and accessible learning experiences, ensures equitable access to educational resources and support, and continuously respond to learner needs in real time. Such ecosystems move beyond static personalization by integrating multiple dimensions of support—cognitive, sensory, and emotional—within a unified framework (Figure 11).

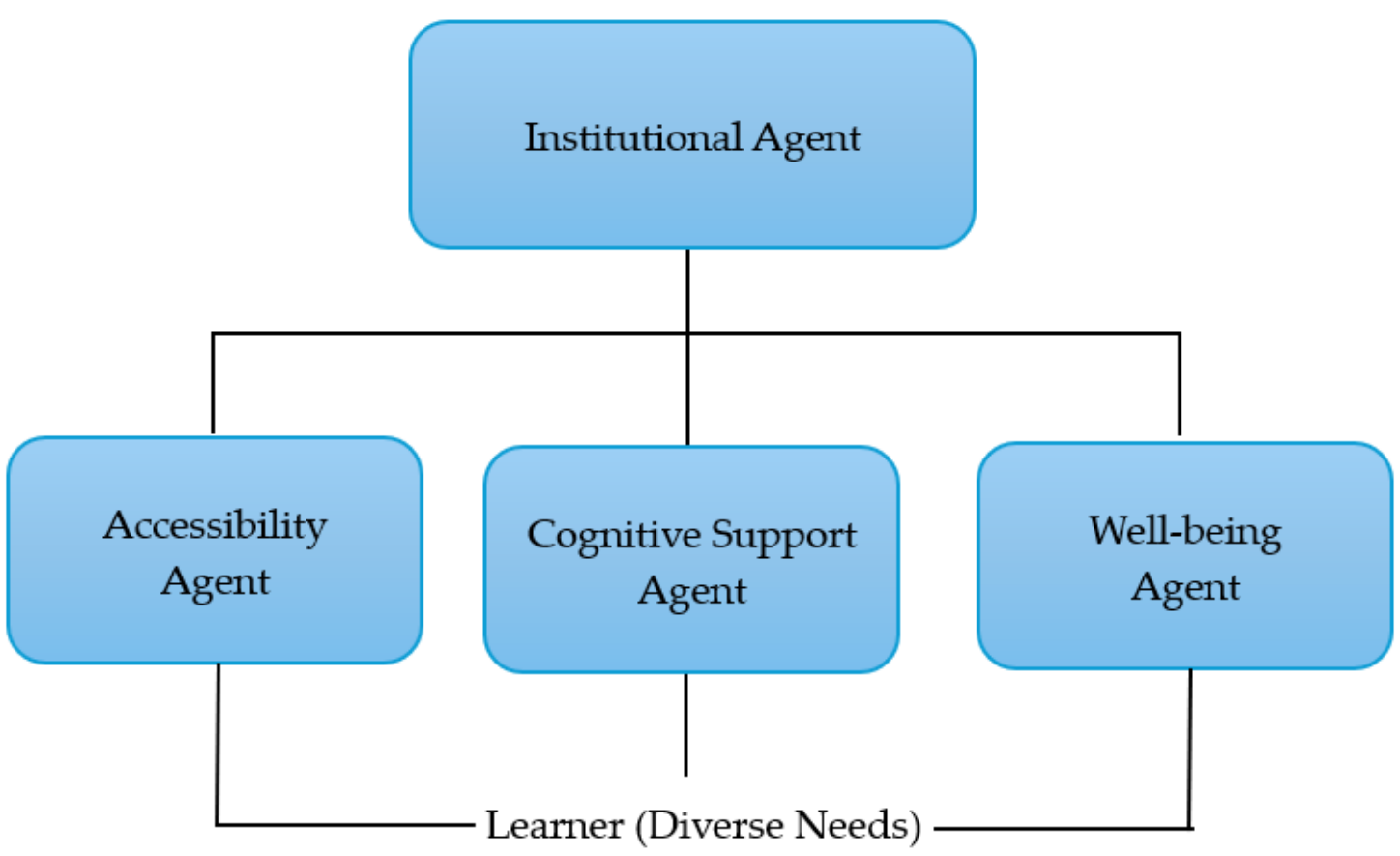

**Figure 11: Conceptual model of an inclusive agentic AI ecosystem illustrating how coordinated multi-agent systems support diverse learners, including students with SENs, through integrated accessibility, cognitive, and wellbeing interventions.**

## 4.2 Supporting Special Educational Needs (SEN)

Agentic AI ecosystems offer a scalable and adaptive approach to supporting students with SEN, encompassing a wide range of learning differences, including cognitive disabilities, neurodiversity, and sensory impairments. By leveraging coordinated multi-agent interactions, these systems enable personalized interventions that align with inclusive education principles (Florian et al., 2011).

### 4.2.1 Cognitive Support

Agentic systems can provide adaptive cognitive support through mechanisms such as personalized pacing, simplified explanations, scaffolding, and reinforcement strategies. These features help address diverse learning needs, including difficulties in comprehension, memory, and attention, thereby promoting deeper understanding and knowledge retention. Such adaptive learning approaches are consistent with research on intelligent tutoring systems and personalized learning environments (Luckin et al., 2016).

### 4.2.2 Sensory Support

For learners with sensory impairments, AI agents can enable accessible and multimodal learning experiences through technologies such as text-to-speech, speech-to-text, visual enhancements, and alternative content representations. These capabilities support inclusive access to educational materials and align with universal design principles in education, ensuring that content is accessible to a broader range of learners (Holmes et al., 2022).

#### 4.2.3 Emotional and Mental Health Support

Agentic AI systems can also support emotional and mental well-being through affect-aware agents capable of detecting indicators of stress, disengagement, or cognitive overload. These agents can provide timely interventions, such as motivational feedback, adaptive task adjustments, or recommendations for support resources. Given the increasing importance of student wellbeing in higher education, such capabilities represent a critical extension of AI-supported learning environments (Inkster et al., 2018; Kasneci et al., 2023).

### 4.3 Multi-Agent Coordination for Inclusion

A defining feature of inclusive agentic ecosystems is the coordination among specialized agents, enabling holistic and context-aware support for diverse learners. Rather than operating as isolated assistive tools, inclusion is achieved through the interaction of multiple agents, including accessibility agents (interface and modality adaptation), cognitive support agents (learning scaffolding and adaptation), wellbeing agents (emotional and engagement monitoring), learning agents (content delivery and assessment). This coordinated approach enables integrated support across multiple dimensions of learning, ensuring that diverse learner needs are addressed simultaneously and dynamically. Such multi-agent coordination aligns with principles of distributed intelligence, where collaborative agent interactions lead to emergent system-level capabilities (Jennings et al., 1998; Wooldridge, 2009).

## 5. Challenges and Considerations

While agentic AI ecosystems present significant opportunities for transforming higher education, their implementation introduces a range of technical, pedagogical, ethical, and inclusion-specific challenges as summarized in Table 1 (Buckingham et al., 2019; Batool et al., 2025; Brohl et al., 2025; Joshi 2025). From a technical perspective, issues such as interoperability, scalability, and data integration highlight the complexity of designing distributed multi-agent systems capable of operating across heterogeneous platforms and large user populations (Wooldridge, 2009). Pedagogically, concerns arise regarding the potential over-reliance on AI, reduced human interaction, and the need to align AI-driven interventions with established learning theories to ensure meaningful educational outcomes (Luckin et al., 2016).

Ethical challenges further emphasize the importance of addressing bias, ensuring data privacy and security, and improving transparency and explainability to foster trust and accountability in AI systems (Russell et al., 2020; Holmes et al., 2022; Acharya et al., 2025; Pinho et al., 2025). In addition, inclusion-specific challenges underscore the risks associated with deploying AI in diverse learning contexts, particularly for students with SEN. Issues such as biased training data, over-automation in sensitive domains, and the necessity for human oversight highlight the importance of adopting inclusive-by-design approaches. Addressing these challenges is essential for ensuring that such systems are effective, trustworthy, and aligned with educational values. Concerns related to bias, transparency, and the ethical use of AI remain central barriers to adoption, particularly in high-stakes educational contexts (Holmes et al., 2022). As AI systems become more autonomous and integrated, these concerns become increasingly complex and require careful consideration.

Collectively, these challenges demonstrate that the successful implementation of agentic AI ecosystems requires not only technological advancement but also careful consideration of pedagogical alignment, ethical responsibility, and inclusive design principles to ensure equitable and human-centered educational environments.

**Table 1. Key challenges in implementing agentic AI ecosystems in higher education, categorized into technical, pedagogical, ethical, and inclusion-specific dimensions.**

| Category | Key Challenges | Description | Implications for Agentic Ecosystems |
|---|---|---|---|
| **Technical** | Interoperability | Integration across heterogeneous systems and platforms | Requires standardised protocols and seamless agent communication |
| | Scalability | Supporting large, diverse user populations | Necessitates robust, distributed multi-agent architectures |
| | Data integration | Managing multimodal, large-scale data | Demands reliable data pipelines and consistency mechanisms |
| **Pedagogical** | Over-reliance on AI | Reduced critical thinking and learner autonomy | Risk of passive learning environments |
| | Reduced human interaction | Diminished student-teacher engagement | Weakens social and collaborative learning |
| | Alignment with learning theories | Lack of pedagogical grounding | Limits educational effectiveness and learning outcomes |
| **Ethical** | Bias and fairness | Algorithmic bias and inequity | May reinforce systemic inequalities |
| | Privacy and Security | Handling sensitive learner data | Requires strict data governance and protection |
| | Transparency & Explainability | Opaque decision-making | Reduces trust and accountability |
| **Inclusion-Specific** | Bias in SEN Contexts | Underrepresentation of diverse learners | Limits accessibility and inclusivity |
| | Over-automation | Inappropriate AI interventions | Risks in sensitive domains (e.g., mental health) |
| | Human Oversight | Need for human-in-the-loop systems | Ensures ethical alignment and contextual relevance |

## 6. Evidence Synthesis

A growing body of research demonstrates the effectiveness of AI in enhancing learning outcomes, student engagement, and personalized learning experiences within higher education. Applications such as intelligent tutoring systems, learning analytics, and adaptive learning platforms have been shown to support individualized learning pathways and improve academic performance (Holmes et al., 2019; Zawacki-Richter et al., 2019). More recently, advances in generative AI have enabled increasingly interactive and context-aware educational support, further strengthening the role of AI in learning environments (Kasneci et al., 2023). Table 2 provides an overview of empirical studies that examine the use and impact of AI agents in undergraduate education. In addition, summary of published studies on Agentic AI and agentic multi-agent AI platform in undergraduate education are briefed in Table 3.

Despite these advances, several limitations persist. Current AI implementations are often fragmented, operating as standalone systems that address specific functions without integration across learning, teaching, and institutional processes. Additionally, challenges related to scalability and interoperability hinder the widespread adoption of AI-driven solutions in complex educational settings. Importantly, there remains a limited focus on inclusive and equitable learning, with many systems insufficiently addressing the needs of diverse learners, including those with special educational needs (Holmes et al., 2022).

These limitations highlight a critical need for integrated, ecosystem-level approaches, where multiple AI components operate in coordination to support holistic and inclusive educational experiences. Agentic AI ecosystems offer a promising pathway to address these gaps by enabling distributed intelligence, continuous adaptation, and coordinated decision-making across educational domains.

### 6.1 Thematic Analysis of AI Agents and Agentic Systems in Higher Education

To complement the evidence synthesis, a thematic analysis was conducted on the studies summarized in Tables 2 and 3 to identify recurring patterns in the development, application, and limitations of AI agents and emerging agentic multi-agent systems in higher education. The analysis followed an inductive coding approach, where each study was examined across four key dimensions: (i) functional roles of AI systems, (ii) system architecture (single agent vs. multi-agent vs. agentic systems), (iii) inclusion and learner support features, and (iv) reported limitations. This approach enabled the identification of cross-cutting themes that characterize the current state of the field and highlight critical research gaps.

A thematic categorization of the studies presented in Tables 2 and 3 reveals a clear distribution of AI system architectures in higher education. Approximately 52% of the studies focus on task-specific AI tools, including intelligent tutoring systems, chatbots, grading systems, and learning analytics platforms. These systems are primarily designed to address isolated educational functions such as content delivery, assessment, or administrative support. Around 28% of the studies employ single-agent AI systems, where a standalone intelligent agent provides interactive support, such as tutoring, advising, or conversational assistance. While these systems demonstrate increased adaptability compared to tool-based approaches, they remain limited in terms of coordination and system-level integration.

In comparison, only about 12% of the studies explore multi-agent systems, where multiple agents collaborate to perform distributed tasks such as tutoring, assessment, and content generation. More notably, agentic multi-agent systems account for approximately 8% of the literature, representing the least explored yet most advanced category. These systems demonstrate capabilities such as role specialization, task orchestration, memory, reasoning, and adaptive coordination across agents. This distribution highlights a significant imbalance in the current research landscape. The dominance of tool-based and single-agent systems indicates that AI in higher education is still largely function-centric and fragmented, focusing on optimizing individual components rather than enabling holistic system intelligence.

In contrast, the relatively limited presence of multi-agent and agentic multi-agent systems suggests that the field has yet to fully realize the potential of distributed, coordinated, and adaptive AI ecosystems. Given the complexity of modern educational environments—characterized by diverse learners, dynamic contexts, and multi-stakeholder interactions—such ecosystem-level approaches are increasingly necessary.

### 6.1.1 Theme 1: Functional Fragmentation of AI Systems

The analysis reveals that a significant proportion of studies in Table 2 focus on function-specific AI applications, such as intelligent tutoring (e.g., AutoTutor, PS2 Pal), automated grading (e.g., Gradescope, Pensieve), conversational agents (e.g., Woebot, Pounce), and learning analytics systems (e.g., Course Signals). While these systems demonstrate measurable improvements in learning outcomes, engagement, and efficiency, they predominantly operate within narrow functional boundaries. Even in more recent implementations, AI systems tend to specialize in discrete tasks such as feedback generation, scheduling, or content delivery, with limited integration across learning processes. This fragmentation persists despite advancements in large language models and multimodal AI systems, indicating that current deployments largely remain task-oriented rather than ecosystem-oriented. In contrast, studies in Table 3 begin to move toward multi-functional coordination, where multiple agents collectively address different aspects of the learning process. However, such approaches are still emerging and not yet widely adopted in real-world educational systems.

### 6.1.2 Theme 2: Evolution from Single-Agent Systems to Agentic Multi-Agent Architectures - Limited Adoption of Multi-Agent and Agentic Systems

A clear progression emerges across the literature from single-agent systems to multi-agent and agentic architectures. Early systems (Table 2), such as pedagogical agents and conversational tutors, are primarily reactive and dialogue-based, with limited autonomy and coordination capabilities. More recent studies (Table 3) demonstrate a shift toward agentic multi-agent systems, characterized by role specialization (e.g., tutor, planner, evaluator, simulator agents), task decomposition and workflow orchestration, memory, reasoning, and self-reflection capabilities, and collaborative and distributed decision-making.

Examples such as GenMentor, MultiTutor, EduVerse, and FACET illustrate how multiple agents can coordinate to support personalized learning, simulate classroom environments, and assist instructors. These systems reflect a transition toward distributed intelligence, where learning processes are managed collaboratively by interconnected agents rather than a single monolithic model. However, despite these advances, many implementations remain at the prototype or conceptual stage, with limited large-scale deployment and empirical validation in authentic educational contexts.

#### 6.1.3 Theme 3: Limited and Fragmented Focus on Inclusive Learning

A critical finding across both tables is the limited and inconsistent integration of inclusive learning principles, particularly for students with SENs. While some studies incorporate elements of personalization, accessibility, or affective computing (e.g., socially assistive agents, EEG-based engagement systems), these features are often isolated rather than systematically integrated. Most AI systems focus on improving general learning outcomes without explicitly addressing cognitive diversity (e.g., learning disabilities, neurodiversity), sensory accessibility (e.g., multimodal interaction support), and emotional and mental well-being.

Even in advanced agentic systems, inclusion is typically implicit rather than explicitly designed, indicating a gap between technological capability and pedagogical inclusivity. Notable exceptions include systems that incorporate affective feedback, multimodal interaction, or adaptive scaffolding; however, these remain limited in scope and scale. This finding strongly supports the need for inclusive agentic AI ecosystems, where accessibility, personalization, and wellbeing support are embedded across coordinated agents rather than treated as add-on features.

#### 6.1.4 Theme 4: Lack of Ecosystem-Level Coordination and Integration

Another dominant theme is the lack of holistic, ecosystem-level coordination across AI systems. While individual applications demonstrate effectiveness, there is limited evidence of integration across learning (student-facing systems), teaching (educator support systems), and institutional processes (administrative and decision-making systems). Even multi-agent systems often operate within bounded domains, such as tutoring, assessment, or simulation, without broader institutional connectivity. This results in siloed intelligence, where insights and capabilities are not shared across the educational ecosystem.

In contrast, several studies in Table 3 (e.g., academic advising systems, classroom simulation environments, workflow-based agent architectures) illustrate the potential for cross-functional coordination, where agents interact across domains. However, such systems are still emerging and lack standardized architectures or widespread implementation.

#### 6.1.5 Theme 5: Persistent Technical, Ethical, and Pedagogical Limitations

Across both tables, several recurring limitations are identified, (i) Technical limitations: **s**calability challenges in multi-agent coordination, dependence on infrastructure and data integration, and limitations of current LLM capabilities (e.g., hallucinations, context constraints), (ii) Ethical and trust-related concerns: bias and fairness in AI decision-making, privacy and data security risks, and lack of transparency and explainability, (iii) Pedagogical constraints: over-reliance on AI systems, limited alignment with learning theories, and insufficient support for long-term learning and self-regulation, and (iv) Adoption and implementation barriers: need for human oversight, instructor readiness and training requirements, and institutional resistance and infrastructure limitations.

These challenges highlight that, despite rapid technological progress, the deployment of AI in education remains complex, context-dependent, and constrained by socio-technical factors.

### 6.2 Synthesis and Implications for Agentic AI Ecosystems

The thematic analysis reveals a clear gap between the current state of AI in education and the potential of agentic AI ecosystems (Figure 12). While existing systems demonstrate strong capabilities in isolated domains, they fall short in delivering integrated, inclusive, and adaptive educational experiences at scale. The findings suggest that future developments should focus on transitioning from function-specific tools to coordinated multi-agent ecosystems, embedding inclusion and accessibility as core design principles, enabling cross-domain interaction between learning, teaching, and institutional agents, and developing scalable, interoperable, and human-centered architectures.

These findings provide strong empirical support for the central argument of this paper: that the future of AI in higher education lies in the development of agentic, multi-agent AI ecosystems capable of delivering holistic, inclusive, and adaptive learning environments. By enabling coordinated interaction among specialized agents, such ecosystems address the limitations of fragmentation and functional isolation observed in the current literature. Importantly, they also provide a scalable foundation for embedding inclusive, personalized, and context-aware learning support, thereby advancing toward more equitable and human-centered educational environments.

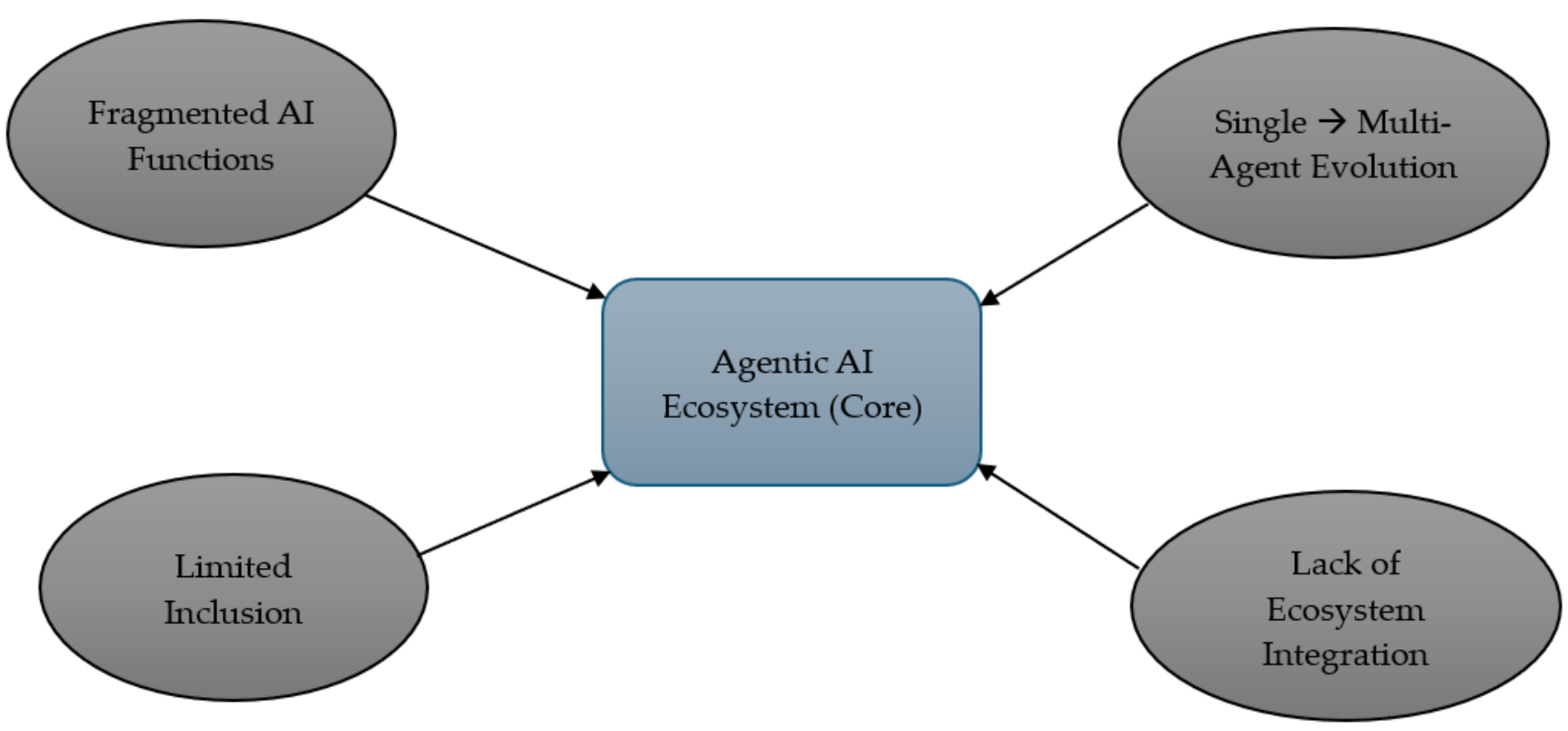

**Figure 12: Thematic map derived from the analysis of AI agents and agentic systems in higher education. The figure illustrates four dominant themes—functional fragmentation, evolution from single to multi-agent systems, limited inclusion, and lack of ecosystem-level integration—converging toward the need for agentic AI ecosystems that enable coordinated, inclusive, and adaptive learning environments.**

## 7. Future Directions

Future research on agentic AI ecosystems in higher education should focus on advancing both technical capabilities and human-centered design principles. Several key directions emerge: (i) Inclusive-by-design AI systems: Embedding accessibility and equity considerations into system architecture from the outset, (ii) Interoperable agent architectures: Developing standardized frameworks to enable seamless integration across platforms and domains, (iii) Human–AI collaboration models: Exploring how humans and AI systems can effectively co-exist and co-evolve in educational environments, and (iv) Real-world implementation studies: Conducting empirical research to evaluate the effectiveness, scalability, and ethical implications of agentic systems in practice. A particularly important direction is the shift toward human–AI co-evolution, where intelligent systems and human stakeholders continuously adapt to each other over time.

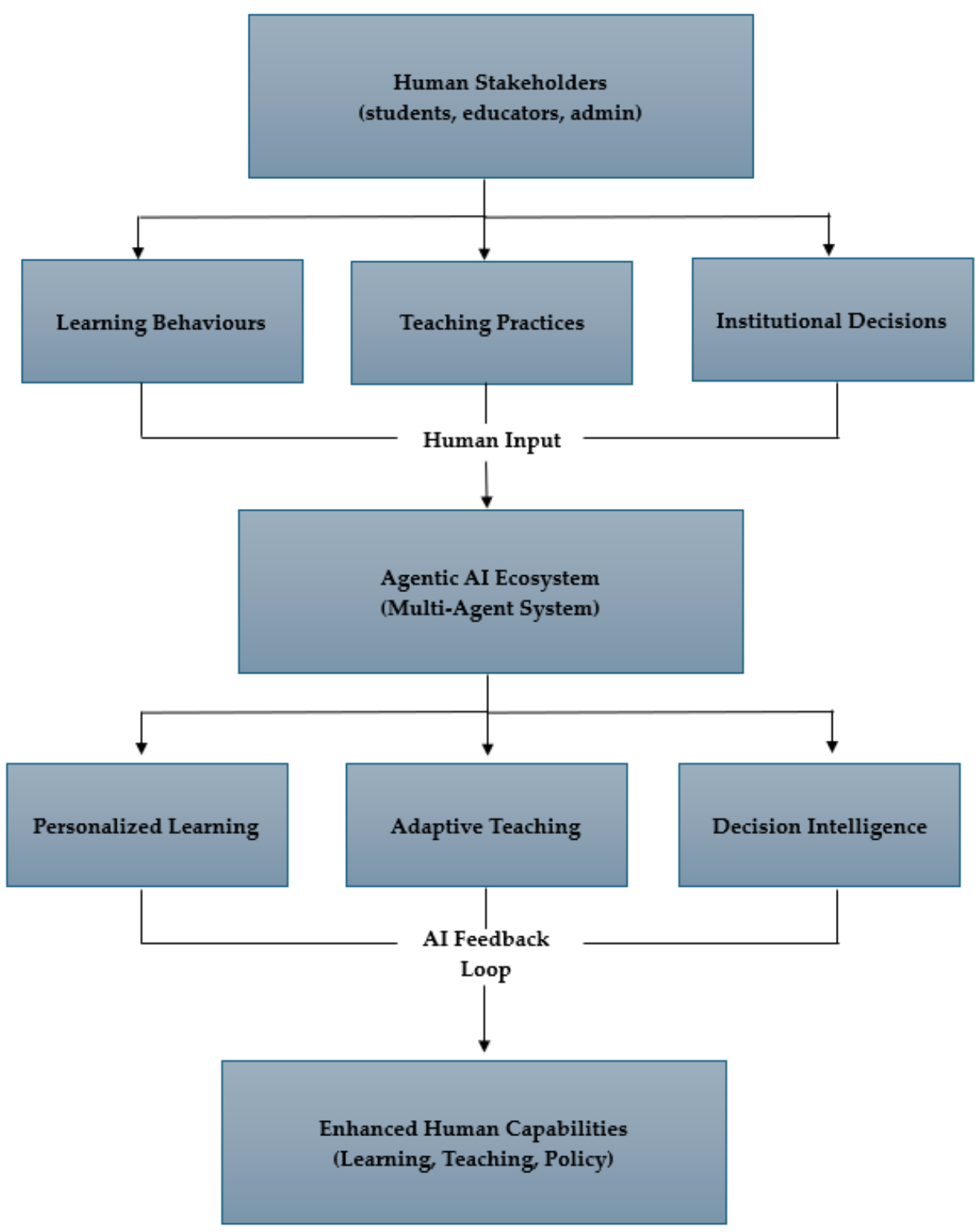

**Figure 13: Human–AI co-evolution model in agentic educational ecosystems, illustrating the bidirectional interaction between human stakeholders and multi-agent AI systems. The model highlights continuous feedback loops through which human behaviors, teaching practices, and institutional decisions inform AI adaptation, while AI-driven personalization, adaptive teaching, and decision intelligence enhance human capabilities, enabling a co-evolving, human-centered educational environment.**

The human–AI co-evolution paradigm represents a critical shift from technology-centric to human-centered AI design in education. As illustrated in Figure 13, agentic AI ecosystems operate through continuous feedback loops in which human inputs—such as learning behaviors, pedagogical practices, and institutional decisions—inform system adaptation. In turn, AI-driven outputs enhance human capabilities by supporting personalized learning, adaptive teaching, and data-informed decision-making. This dynamic interaction fosters a co-evolving ecosystem in which both human and artificial agents continuously learn, adapt, and improve. Such an approach reinforces the importance of human-in-the-loop design, ensuring that AI systems remain aligned with pedagogical goals, ethical principles, and inclusive educational practices (Luckin et al., 2016; Holmes et al., 2022). Figure 14 illustrates a graphical overview of future agentic multi-agent AI platform for education.

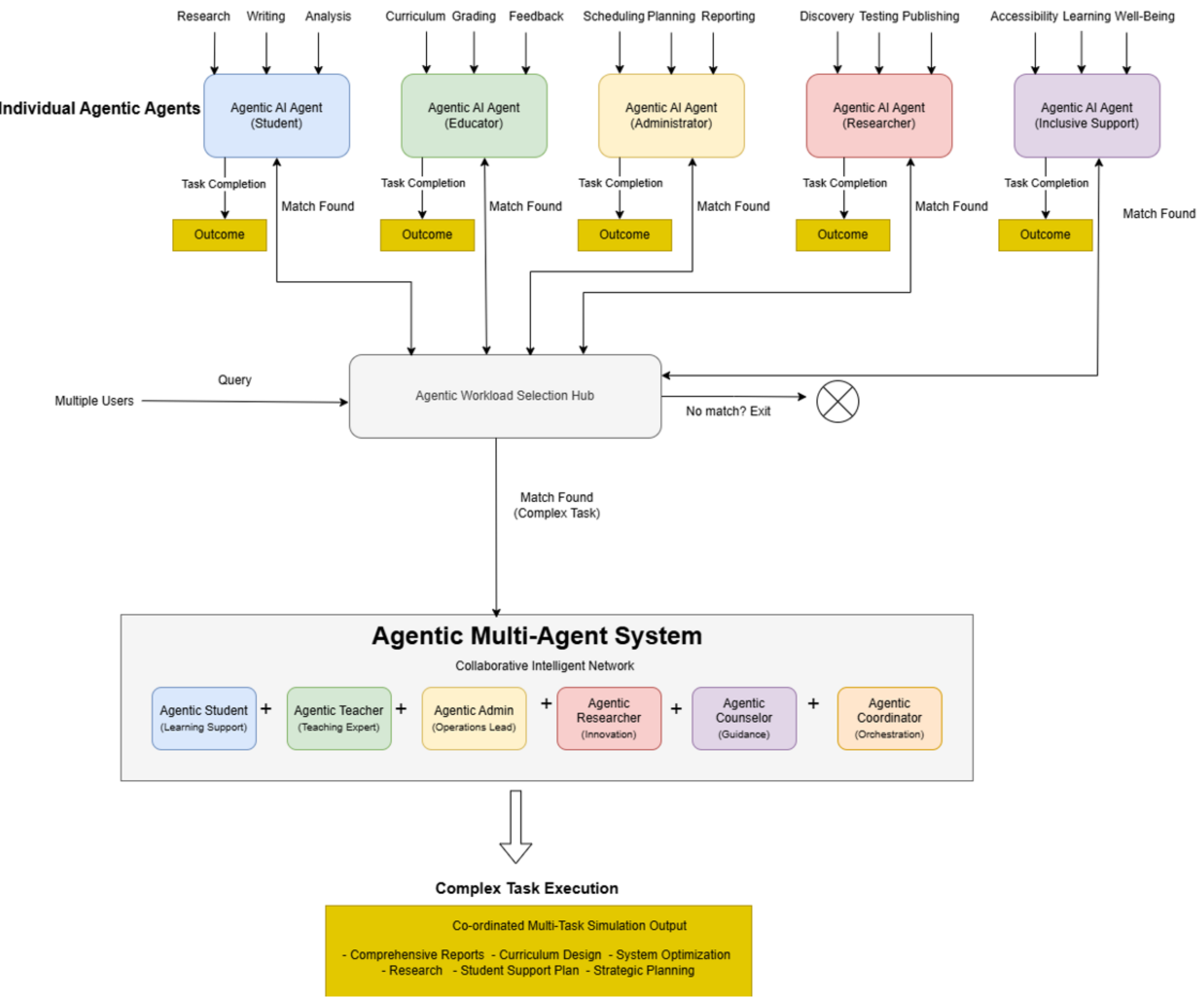

**Figure 14: Illustration of Future Agentic multi-Agent AI Ecosystem**

## 8. Conclusion

This paper presents a forward-looking perspective on the evolution of AI in higher education toward agentic AI ecosystems. By enabling the integrating of learning, teaching, and institutional processes within a unified and coordinated framework, these systems offer the potential to fundamentally transform educational environments. Importantly, this transformation extends beyond efficiency and automation to emphasize inclusive and equitable learning, ensuring that diverse learner needs—including those of students with SENs—are addressed in a scalable and adaptive manner. Realizing this vision requires sustained interdisciplinary collaboration, robust ethical and governance frameworks, and a strong commitment to human-centered design principles. Rather than replacing human roles, agentic AI ecosystems are envisioned to augment and co-evolve with students, educators, and institutions through continuous feedback and adaptive interaction. Ultimately, the future of higher education will not be defined by isolated, task-specific AI tools, but by intelligent, inclusive, adaptive, and coordinated interconnected agentic ecosystems capable of delivering personalized, context-aware, and equitable learning experiences at scale.

**Table 2: Summary of published study conducted on the AI agents (single/multi-agents) and its impact in undergraduate education**

| Author (Year) | Stakeholders | AI Agents | Purpose | Limitations/Future Work |
|---|---|---|---|---|
| Graesser AC et al., (2001) | Students, Educators | AUTOTUTOR, ANDES, ATLAS, WHY2 | Tutoring computer literacy, physics conceptual knowledge | extend to additional dialogue types; handle complex student language |
| Pantic M et al., (2003) | Students | Simple Agent Framework (SAF) | Teaching AI basics | unresponsiveness under infinite loops, need for more robust error handling, GUI improvements |
| Graesser AC et al., (2005) | Students, Instructors | AutoTutor | Simulate human tutoring via natural language dialogue to improve STEM learning | Emotion-sensitive dialogue and 3-D interactive simulation evaluation still in progress |
| Pantic M et al., (2005a) | Students | SAF | Teaching introductory AI programming, AI concepts | partial scalability for novices, need for extending SAF to more complex AI courses |
| Pantic M et al., (2005b) | Students | Fleeble Agent Framework | improving student engagement, comprehension on programming | Need to involve wider use in AI courses, improved pedagogical tools |
| Biswas G et al., (2005) | Students | Teachable agent "Betty" with qualitative reasoning, concept maps, mentor agent "Mr. Davis" with feedback and hints | Enhance student learning, understanding, self-regulation, reflection, transfer of knowledge | passive and quiz-focused; need - adding self-regulation strategies, interactive agent behaviors, extended reasoning capabilities |
| Baylor A et al., (2005) | students, pre-service teacher | pedagogical agents: Expert, Motivator, Mentor | motivation, learning, self-efficacy, instructional planning transfer | short interaction time and scripted agent behavior, need to include longer interaction studies |
| D'Mello S et al., (2012) | students, educators | AutoTutor | engagement, confusion, frustration, boredom | Individual differences not fully addressed, need for models on broader populations and contexts |
| Arnold KE et al., (2012) | Students, instructors | Course Signals (learning analytics early alert system) | Predicting risk students | managing student message overload; future expansion planned to larger student populations |
| Lee H et al., 2015 | students | Pedagogical Agent with Gaze Interaction (PAGI) | Word learning test: Korean vocabulary to Japanese students | simple learning material (basic vocabulary), future work suggests testing motivation effects via fixation duration |
| Hooshyar D et al., 2015 | Students, experienced tutor | Multi-agent system in FITS | Problem-solving ability, personalized guidance, adaptive navigation, boost programming skills | Limited to basic C concepts, scope restricted to novice level; need to expand concepts |
| Fitzpatrick KK et al., (2017) | Students with anxiety/depression | Woebot | depression, anxiety, emotional awareness, user engagement, satisfaction | Small sample size; short intervention (2 weeks); no follow-up data; no mediator analysis; lack of engagement data for control group; suggested replication with larger samples, longer dose, follow-up |
| Singh A et al., (2017) | Educators, students | Gradescope | Speed, consistency, flexibility in grading handwritten homework and exams | limited automation in grading itself; need work on concept tagging and longitudinal student performance analysis |
| Page LC et al., (2017) | students | Pounce | Reduce summer melt, personalized text message–based outreach | Initial human supervision required, need studies to examine impacts on persistence and degree completion |
| Goel AK et al., (2018) | students, educators | Jill Watson | Reduce TA load by answering FAQs, welcoming students; sustain engagement and support scaling in large online courses | Ethical concerns on transparency; limited AI capabilities in nuanced queries; further research on effectiveness, transferability, and ethics |
| Barrett M et al., (2019) | Students, faculty, administrati ve staff | Conversational AI chatbots (e.g., Pounce, ACE), contextual user interfaces | Enhance student services, increase enrolment and retention, automate routine administrative and advising functions | Limited AI depth in contextual understanding; infrastructure and cost challenges; ongoing need for human oversight and improvements in AI responsiveness |
| Hsu MH et al., 2021 | Students, educators | TPBOT (TOEIC Practice Chatbot): task-oriented AI chatbot | reduce foreign language anxiety, and boost oral proficiency via interactive, corrective dialogues | Variable practice frequency hard to control; limited to TOEIC speaking; need work to integrate with personalized recommender systems for broader EFL support |
| Cristina DBM et al., (2022) | Students, administrati ve staff, educators | Chatbot designed for Stricto Sensu Postgraduate Secretariat inquiries | Alleviate administrative workload; reduce student waiting times; improve response efficiency and satisfaction | Initial deployment limited by pandemic conditions; maintaining humanized service valued; further refinement and expansion planned |
| Meyer K et al., (2023) | students, educators | Pounce | Impact on course grades, assignment completion, course retention, and student engagement | Requires substantial upfront effort from instructors; ongoing human oversight needed; need work on scaling to other courses and modalities |
| Baillifard A et al., 2023 | Students, instructor | Personal AI tutor app | personalization, spaced repetition, retrieval - to boost long-term retention and performance | No randomized active control; self-selected usage; need - test personalization ablation, identify optimal behaviors |
| Ng DTK et al., (2023) | educators, students | Various AI-driven educational tools and intelligent tutoring systems | AI digital competencies to enhance teaching, learning, assessment, and student AI literacy | Conceptual framework, lacks empirical validation; need work on empirical testing and program design |
| Chen Y et al., (2023) | Students, Educators | Conversational chatbot "Sammy" | teaching AI concepts, improve student engagement, responsiveness, interactivity, confidentiality, exploration of resources | Challenges in conversational flow, limited emotional connection, difficulty understanding nuanced language and slang, student superficial engagement; need work on enhancing chatbot IQ and EQ, improving conversational abilities, expanding linguistic understanding, addressing ethical AI principles like transparency, fairness, privacy, and trust |
| Bran AM et al., (2023) | Chemists, non-experts | ChemCrow | Autonomous synthesis, novel chromophore discovery, expert | reasoning flaws/hallucinations; reproducibility via closed APIs; need to expand tools (language/image), diverse benchmarks, open-source models |
| | | | evaluations, automate chemical reasoning/planning/execution | |
| Taneja L et al., (2024) | Students, educators | Jill Watson | student support - answering course-related queries, improved engagement and responsiveness | Difficulty in AI understanding nuanced student questions, need for ongoing human oversight and ethical considerations |
| Neupane S et al., (2024) | students, faculty, staff | BARKPLUG V.2 | contextualized, accurate responses to diverse university resource queries | Lacks automatic speech recognition and multilingual support; occasional hallucinations; token limit hinders long conversations; need work on language support |
| Ounejjar LA et al., (2024) | Students, educators, academic administrators | SmartBlendEd platform | Efficient scheduling, personalized timetables, enhanced student engagement and learning outcomes | data privacy concerns; need for ongoing algorithm refinement, broader feature expansion, and user training |
| Xu J et al., (2024) | teachers, students | Inter-Agent system | interdisciplinary teaching designs; enhanced clarity, thematic focus, integration of disciplines; refined and coherent lesson plans | need to expand knowledge base and develop interactive human-machine design interfaces; incorporate meta-learning and reinforcement learning for innovation |
| Mithun P et al., (2024) | Students, Educators, Institutional Staff | Adaptive scheduling algorithm, pseudorandom algorithm for auto-scheduling | Efficient class scheduling, communication via news sharing, timely reminders, centralized resource access, student engagement | Scalability issues with more users, user acceptance needed, integration with existing systems needed, limited real-world testing; need work for better AI integration, expanded collaboration tools, enhanced security, more customization, learning analytics integration |
| Dieker LA et al., (2024) | Teachers, Students (especially with disabilities) | Socially assistive AI agent "ZB" and robot "Ray-Z" | Support teacher instruction, improve student engagement, communication, social-emotional regulation, coding skills, and self-regulation; personalized learning support for neurodiverse students | Technical barriers, AI bias, need for teacher training, implementation fidelity challenges, complexity of multimodal emotion recognition, and privacy concerns; future focus on accessibility, enhanced robustness, expanded AI-agent customization, and integration in teacher education programs |
| Dai CP et al., (2024) | Students, teachers | conversational agents (CAs) named Evelyn | authentic teaching simulation, enhance pedagogical reasoning, improve reflective teaching practices, engagement, and social interaction through humor | use of GPT-2 (not the latest AI), exploratory nature, limited conversation variety; need work for improving agent dialogue diversity, integrating newer AI models, and expanding design for HCAI |
| Mollick E et al., (2024) | Students, Instructors | Multiple AI agents: Mentor, Investor, Evaluator, Progress, Class Insights | Facilitate personalized simulated practice, student engagement, adaptive feedback, instructor insights into student learning progress | AI consistency and bias, context limitations, potential inaccuracies (hallucinations), adapting AI for specific domains, ensuring ethics and transparency, ongoing pedagogical evaluation, and improving multi-agent coordination and simulation quality |
| Ahmed Z et al., (2024) | students | ChatGPT | teaching assistant for introductory programming | Limitations in resolving complex confusions, individualized responses, tailored debugging; work need for broader evaluation, integration guidelines, and human-AI collaboration |
| Power R (2024) | students, educators | ChatGPT and similar AI agents | student engagement, instructional strategies, and classroom management | Small sample size, lack of control group; need work on expanded pedagogical training for AI integration, and larger studies for reliability |
| Darvishi A et al., (2024) | students | AI support within the RiPPLE peer review platform | AI assistance and self-regulation on student agency and the standard of peer feedback | Short duration (8 weeks), comparison groups received AI initially; need qualitative insight, better hybrid human-AI design, address AI biases, explore other learning tasks beyond peer review |
| Kestin G (2024) | Students, instructors | PS2 Pal | personalized, self-paced tutoring | Ceiling effects underestimate gains; future: full courses, spacing, homework/labs, multimodal feedback |
| Yang Y et al., (2025) | Instructors, students | Pensieve Grader | End-to-end AI-assisted grading | Performance depends on rubric quality, response clarity, and domain; still requires human-in-the-loop review; focuses on STEM and free-form answers, leaving broader subjects |
| Hao Z et al., (2025) | students, educators, administrato rs | MAIC multi-agent system | learning, motivation, interaction between students and multiple AI agents | Study limited to one elite university course; short-term effects only; unclear long-term retention, need work on optimized agent designs, contexts, and diverse populations |
| Baradari D et al., 2025 | students; educators | NeuroChat | engagement and, learning | Short sessions (20min), no long-term learning gains; high inter-subject EEG variability; future: multimodal, longitudinal, user-configurable styles/profiles |
| Jin L et al., (2025) | students, educators | Chat Melody | Learning gains in music theory; cognitive load reduction during music analysis tasks | Small sample size; limited to short-term outcomes; need larger, diverse groups and long-term retention studies |
| Sajja R et al., (2025) | students, educators | Educational AI Hub | Student engagement, perceptions of trust, ethics, usability, learning outcomes; | Ethical ambiguity and academic integrity concerns; need clearer institutional policies and faculty guidance; explore long-term learning and self-regulation impacts |
| Wang H et al., (2025) | students, educators | LLM based AI-Agent system for collaborative coding | Learning achievement, self-efficacy, mental effort, engagement levels | Small sample size; risks of AI over-dependence; AI-Agent limited by current LLM capabilities; need work to integrate multimodal learning, knowledge graphs, causal reasoning |
| Cayli O et al., (2025) | institutions, administrati ve staff, students | LMSOPT system for demand forecasting | Optimize classroom scheduling and resource use; balance conflicting objectives like occupancy rate, waiting time, teacher availability; improve operational efficiency and student satisfaction | pilot needed for long-term validation; potential extensions with ensemble models and reinforcement learning; integration of NLP for feedback analysis |
| Jain RK et al., (2025) | Students, educators, administrators | ClassSync platform | Automate attendance, personalized learning recommendations, real-time dashboards and analytics; classroom management efficiency and student engagement | high development cost, data privacy challenges, system complexity, user adaptation barriers; plans for blockchain certification, VR learning, AI proctoring, multilingual support |
| Kestin G et al., (2025) | students, educators | AI tutor "PS2 Pal" | engagement, motivation, in-class active learning | AI hallucinations; limited to STEM topics; need work to include multimodal interactions, broader contexts, and further pedagogical integration |
| Simmhan Y et al., (2025) | students, educators | AI Instructor Agent integrated with Microsoft Teams Copilot | engagement evaluation; support inquiry-driven, self-paced learning; personalized feedback | Small pilot with 17 students; occasional AI hallucinations; need to correlate engagement metrics with learning outcomes; need work to integrate Agentic AI with tool use and improve evaluation |
| Yan L et al., (2025) | students | Generative AI agents | scaffolding and data storytelling | 1-hour intervention duration limits long-term retention assessment; need work to explore longer interventions, diverse samples, and detailed design features impact |
| Kshetri N et al., (2025) | Students, educators, administrative staff, university management | Various AI agents and agentic AI systems: Cogniti platform, CollegeVine AI Advisor | Transform teaching, personalized tutoring; streamline administration task; increase student engagement; improve recruitment efficiency | Privacy and cybersecurity risks; transparency and explainability challenges; risk of job displacement; requirement for human oversight; evolving regulatory and governance frameworks needed |
| Wang F et al., (2025) | students, educators | AI-based interactive scaffolding system | Speaking performance, goal setting, self-evaluation, motivation in learning | Limited to informal digital learning; potential variability in implementation fidelity; need work on scalability and diverse educational contexts |
| Chen S et al., (2025) | instructors, pedagogy experts | LLM-powered interactive pedagogical agent chatbot | instructor AI adoption; pedagogically relevant teachings; support AI-conservative and novice instructors | Limited instructor-AI conversational data; need to incorporate diverse pedagogical resources and human-in-the-loop refinement; assess long-term impacts on adoption |
| Marietta OO et al., (2025) | administrative staff, educators | AI-powered application tools | administrative efficiency, automate routine tasks, enable focus on strategic activities, enhance skill development | inadequate infrastructure, resistance to change, lack of training, and support; future strategies emphasize hands-on training, infrastructure investment, and cultural shifts |
| Patel CD et al., (2025) | Students, educators, administrators | Intelligent Tutoring Systems: Pedagogical, Conversational, Adaptive Assessment, Automated Feedback and Grading, Affective Computing Agents | personalized learning, automated feedback, engagement, and administrative support; | ethical concerns (privacy, bias), high development costs, need for human intervention; future: affective computing, collaborative learning, transparency, long-term impact |
| Di Z et al., (2025a) | students, educators | AI Agents | competency development across learning ability, planning, execution, expression, collaboration | AI over-dependence, privacy concerns, field adaptability; need work on building adaptive, self-evolving AI agents and human-AI hybrid training |
| Di Z et al., (2025b) | students | DeepSeek-based multi-agent system | for graduate training in coursework, research, and practice | Currently theoretical and framework-level; no quantitative learning outcomes yet; future: implement and refine in real programs |
| Jiang W et al., (2025) | Students | Academic Info Extractor, Query Rewriter, Query Classifier, Thought Agent, Action Agent, Answer Generator, Quality Controller | Automates info retrieval/drafting for CS graduate advising with human-in-loop | Small user study (8 CS advisors); no student/admin views; lacks real deployment; future: diverse/multi-institutional studies, longitudinal testing, student perspectives, adaptive formatting |
| Rogers K et al., (2025) | students; instructors | MatlabTutee, MatlabGPT | active interactions (explaining/reasoning); to reveal knowledge gaps | Low voluntary adoption; future: better feedback without demotivation, curriculum integration |

**Table 3: Summary of work showing the current available Agentic AI and possibilities for Agentic multi-agent AI in undergraduate education**

| Author (Year) | Stakeholders | AI Agents | Purpose | Agentic Type |
|---|---|---|---|---|
| Abdelhamid AA et al., 2021 | Students, Advisors | Six agents: Student, Instructor, Administrator, Schedule, Performance Monitoring, Smart Advisor Agents | Advising, tracking progress, suggestions, scheduling | Agentic multi-agent |
| Jiang YH et al., 2024 | Students, Instructors | LLM-based AI Agents | human learners supporting, construction strengthening of knowledge | Agentic multi-agent |
| Wei H et al., 2024 | Medical students | Agentic patient, medical expert, radiologist, medical student | medical education: simulation | Agentic multi-agent |
| Zha S et al., 2024 | Students | Mentigo | creative problem-solving, giving feedback and scaffolding | Agentic |
| Thway M et al., 2024 | Students | Professor Leodar (RAG chatbot) | personalized guidance, 24/7 availability, and contextually relevant responses. | Agentic |
| Degen PB et al., 2025 | Students | Socratic AI Tutor | dialogic scaffolding | Agentic |
| Gupta R et al., 2025 | Instructors | CrewAI framework with 4 specialized agents: Curriculum Planning, Content Generation, Linguistic Adaptation, Assessment Synthesis | generating culturally adaptive educational content in African languages. | Agentic multi-agent |
| Kamalov F et al., 2025 | Students | Multi-agent system for automated essay scoring (MASS), other examples: PitchQuest agents (Mentor, Investor, Evaluator), MEDCO agents (patient, doctor, radiologist) | Essay writing assessment. | Agentic multi-agent |
| Zeng B et al., 2025 | Students | Diagnostician, Planner, Tutor Agents (LangGraph-orchestrated) | For customised learning | Agentic multi-agent |
| Gonnermann MJ et al., 2025 | Instructors | FACET multi-agent system: Learner, Teacher, Evaluator agents | student profiles simulation, building worksheet contents, quality control | Agentic multi-agent |
| Wang T et al., 2025 | Students | GenMentor multi-agent system | Skills identification, crafting personalized content. | Agentic multi-agent |
| Sun E et al., 2025 | Students | MultiTutor agents: Manager, Content Expert, Scaffolding, Visual Learning, Resource Recommender, Researcher | Multimodal (images, animations, practice problems, simulations) tutoring | Agentic multi-agent |
| Zhang Z et al., 2025 | Instructors | SimClass agents: Teacher, Assistant, Classmate Agents | Classroom dynamics simulations | Agentic multi-agent |
| Shi Y et al., 2025 | Students, Instructors | EducationQ triadic multi-agent framework | simulate realistic teaching interactions or mimic more realistic teacher-student-evaluator dynamics | Agentic multi-agent |
| Hajji MEI et al., 2025 | Students | LLM-powered NPC tutor | Tutoring using multimodal channels (voice, gaze, gesture) | Agentic |
| Zhao TF et al., 2025 | Students | InqEduAgent | simulates learning partners for inquiry-based learning | Agentic multi-agent |
| Ma Y et al., 2025 | Students, Instructors | EduVerse multi-agent system | simulate classroom dynamics<br>Includes cognition, interaction, emotion | Agentic multi-agent |
| Nguyen DQ et al., 2025 | Students | V-MATH multi-agent system: Planner, Executor agents with Memento memory mechanism | question generation, solving, and personalized tutoring | Agentic multi-agent |
| Jiang YH et al., 2025 | Instructors, University Students | AWE (Agentic Workflow for Education) platform | educational workflow: self-reflection, tool calling, job planning | Agentic multi-agent |
| Wang J et al., 2025 | Students | TRAVER agent workflow | tutoring with feedback / evaluation | Agentic |
| Rendy AR, 2025 | Students | Intelli Tutor | multimodal personalized tutoring offering dynamic, step-by-step guidance | Agentic |
| Zhang X et al., 2025 | Students | EduPlanner: Evaluator, Optimizer, Analyst Agents | Generate, evaluate and optimize customized instructional designs for math lessons | Agentic multi-agent |

**Disclosure of AI-Assisted Technologies and Generative AI in the Writing Process**

In preparing this work, authors utilized ChatGPT as a language editing aid. All AI-generated suggestions were subsequently reviewed and revised by the authors, who bear full responsibility for the published material.